%% file: colm2026_conference.tex
\documentclass{article} 
\usepackage[preprint]{colm2026_conference}

\usepackage{microtype}
\usepackage{hyperref}
\usepackage{url}
\usepackage{booktabs}
\usepackage{amsmath,amssymb}
\usepackage{graphicx}
\usepackage{algorithm}
\usepackage{algorithmic}
\usepackage{subcaption}
\usepackage{xcolor}
\usepackage{pifont}
\usepackage{tikz}
\usetikzlibrary{calc,decorations.pathreplacing,arrows.meta,positioning,fit,backgrounds}
\usepackage{lineno}

\usepackage{enumitem}
\usepackage{placeins}
\usepackage{wrapfig}
\newenvironment{denseitemize}{
\begin{itemize}[topsep=2.5pt, partopsep=0pt, leftmargin=1.5em]
  \setlength{\itemsep}{2.5pt}
  \setlength{\parskip}{0pt}
  \setlength{\parsep}{0pt}
}{\end{itemize}}

\usepackage{tikz}
\newcommand{\blackcircled}[1]{%
    \tikz[baseline=(char.base)]{
        \node[shape=circle, fill=black, draw=black,
              inner sep=1.5pt, text=white,
              font=\footnotesize\bfseries] 
              (char) {#1};
    }%
}

\usepackage{xcolor}
\usepackage{multirow}
\definecolor{themecolor}{HTML}{FF5F03}
\definecolor{darkblue}{rgb}{0, 0, 0.5}
\hypersetup{colorlinks=true, citecolor=darkblue, linkcolor=darkblue, urlcolor=darkblue}

\input{math_commands.tex}

\title{Introspective Diffusion Language Models}

\author{
Yifan Yu$^{*,1,2}$, Yuqing Jian$^{*,1}$, Junxiong Wang$^{1}$, Zhongzhu Zhou$^{1}$, \\
Donglin Zhuang$^{1}$, Xinyu Fang$^{1}$, Sri Yanamandra$^{1}$, Qingyang Wu$^{1}$, Tri Dao$^{1,4}$ \\ Xiaoxia Wu$^{1}$, Shuaiwen Leon Song$^{1}$, Ben Athiwaratkun$^{1}$, \\
James Zou$^{\dagger,1,5}$, Fan Lai$^{\dagger,\diamondsuit,2}$, Chenfeng Xu$^{\dagger,\diamondsuit,1,3}$ \\
[0.5em]
$^{1}$Together AI \quad
$^{2}$University of Illinois Urbana-Champaign \quad
$^{3}$The University of Texas at Austin \\
$^{4}$Princeton University \quad
$^{5}$Stanford University \\
[0.3em]
{\small $^{*}$Equal contribution \quad
$^{\dagger}$Equal advising \quad
$^{\diamondsuit}$Corresponding author}
}

\begin{document}

\ifcolmsubmission
\linenumbers
\fi

\maketitle
\begin{abstract}
Diffusion language models promise parallel generation, yet still lag behind autoregressive (AR) models in quality. We stem this gap to a failure of introspective consistency: AR models agree with their own generations, while DLMs often do not. We define the introspective acceptance rate, which measures whether a model accepts its previously generated tokens. This reveals why AR training has a structural advantage: causal masking and logit shifting implicitly enforce introspective consistency. Motivated by this observation, we introduce Introspective Diffusion Language Model (I-DLM), a paradigm that retains diffusion-style parallel decoding while inheriting the introspective consistency of AR training. I-DLM uses a novel introspective strided decoding (ISD) algorithm, which enables the model to verify previously generated tokens while advancing new ones in the same forward pass. 
From a systems standpoint, we build I-DLM inference engine on AR-inherited optimizations and further customize it with a stationary-batch scheduler. To the best of our knowledge, I-DLM is the first DLM to match the quality of its same-scale AR counterpart while outperforming prior DLMs in both model quality and practical serving efficiency across 15 benchmarks. It reaches 69.6 on AIME-24 and 45.7 on LiveCodeBench-v6, exceeding LLaDA-2.1-mini (16B) by more than 26 and 15 points, respectively. Beyond quality, I-DLM is designed for the growing demand of large-concurrency serving, delivering about 3× higher throughput than prior state-of-the-art DLMs.


\end{abstract}

\vspace{-0.2cm}
\begin{center}
\large
\href{https://github.com/Introspective-Diffusion/I-DLM}{\textbf{Code}} \quad $\vert$ \quad
\href{https://introspective-diffusion.github.io}{\textbf{Website}} \quad $\vert$ \quad
\href{https://huggingface.co/collections/yifanyu/introspective-diffusion-language-models-i-dlm}{\textbf{Models}}
\end{center}
\vspace{-0.2cm}

\input{sections/introduction}
\input{sections/motivation}
\input{sections/method}

\input{sections/experiments}
\input{sections/conclusion}


\bibliography{colm2026_conference}
\bibliographystyle{colm2026_conference}

\input{sections/appendix}
\end{document}

%% file: math_commands.tex
\usepackage{amsmath,amsfonts,bm}

\def\eqref#1{equation~\ref{#1}}

\def\1{\bm{1}}

\DeclareMathAlphabet{\mathsfit}{\encodingdefault}{\sfdefault}{m}{sl}
\SetMathAlphabet{\mathsfit}{bold}{\encodingdefault}{\sfdefault}{bx}{n}



%% file: sections/introduction.tex
\section{Introduction}
\label{sec:intro}

\begin{figure*}[t]
\centering
\includegraphics[width=.97\textwidth]{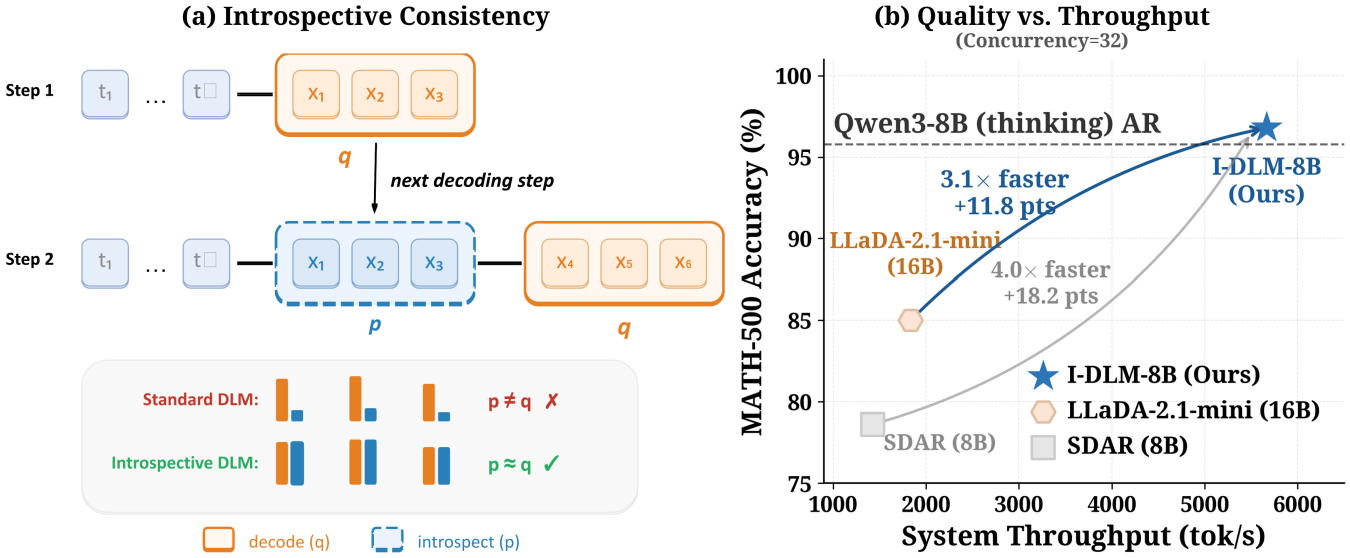}
\caption{\textbf{(a)} Introspective consistency: standard DLMs generate tokens whose distributions $q$ diverge from the model's own next-step predictions $p$; I-DLM trains generation and introspection to agree ($p \approx q$). \textbf{(b)} Quality vs.\ throughput on MATH-500: I-DLM-8B matches Qwen3-8B (thinking) AR performance while achieving 3.1$\times$ higher throughput and +11.8 points over LLaDA-2.1-mini (16B), and 4.0$\times$ higher throughput over SDAR (8B).}
\label{fig:teaser}
\vspace{-.6cm}
\end{figure*}

Diffusion language models (DLMs)~\citep{austin2021structured,sahoo2024simple,nie2025large,cheng2025sdar} offer an appealing alternative to autoregressive (AR) language models: by iteratively refining a block of tokens, they break the sequential bottleneck of next-token decoding and enable parallel generation. Yet this promise remains largely unrealized, as shown in Figure~\ref{fig:teaser}. First, a substantial quality gap between DLMs and AR models remains a key barrier to adoption (\S\ref{sec:motivation}). Second, from an efficiency standpoint, limited system support for diffusion inference prevents DLMs from translating their theoretical parallelism into practical speedups.

The historical trajectory of DLM development is highly revealing. Across a broad line of work, from early continuous formulations~\citep{li2022diffusionlm} and uniform-state DLMs~\citep{austin2023structureddenoisingdiffusionmodels} to discrete diffusion models~\citep{lou2024discretediffusionmodelingestimating}; and from fully bidirectional attention~\citep{nie2025largelanguagediffusionmodels}, to blockwise decoding~\citep{cheng2025sdar,wu2025fast} and causal-mask decoding~\citep{liu2025wedlm}, each generation moves closer to discrete AR language models. Increasingly AR-like training signals have also been introduced to improve optimization and quality~\citep{liu2025tidar,gat2025setblockdecodinglanguage,ye2025dream,tian2025next}. In other words, much of the field has implicitly converged on the same intuition: the path to stronger DLMs should progressively move closer to AR models.
In this work, we argue for a different  trajectory. Rather than beginning from diffusion and asking how to make it more like AR, we begin from AR and ask: \emph{what is the essential principle that makes AR models so strong, and can it be preserved in a parallel generation paradigm?}

We revisit this question from two perspectives: algorithms and systems. From the algorithmic perspective, this change in viewpoint leads us to re-examine a deceptively simple design at the core of AR modeling: causal masking together with logit shifting. Beyond enabling next-token prediction, this training structure implicitly teaches the model to revisit and validate its previously generated tokens under the same predictive rule. In effect, AR models are trained to agree with what they generate, exposing a fundamental limitation of existing DLMs: although they can generate tokens in parallel, they are generally \emph{not} trained to agree with their own generations (e.g., due to multi-step bidirectional denoising). We formalize this gap through the notion of \emph{introspective acceptance rate}, which measures whether a model internally accepts its previously generated tokens. We find that this property serves as a useful proxy for the degree to which a DLM remains consistent with its own generations.

From the systems perspective, existing DLMs are often optimized for aggressive low-latency decoding, but this comes at the cost of substantially higher computational overhead. While such overhead can be partially hidden in memory-bound regimes, production deployments requiring large-batch inference quickly hit the compute-bound regime. Unfortunately, LLM serving stacks are poorly matched to the multiple-query, multiple-denoising patterns. 

\textbf{Contributions.}  These algorithmic-system mismatches motivate us to co-design the \emph{Introspective Diffusion Language Model} (I-DLM), a new paradigm that preserves the introspective consistency of AR while retaining parallelism.
I-DLM is built on introspective-consistency training, an efficient recipe for converting pretrained AR models into DLMs, and a novel introspective strided decoding (ISD) algorithm that generates N tokens per forward pass while verifying prior tokens against a causal anchor distribution. We show that explicitly enforcing introspective consistency during training is key to substantially closing the quality gap between DLMs and strong same-scale AR models. 
We make the following contributions: 
\begin{denseitemize}
   \item \textbf{A key insight: introspective consistency is the missing principle in prior DLMs.}
We show that diffusion language models do not inherit the \emph{introspective consistency} of AR models: they are not trained to agree with their own generations. This missing property fundamentally limits their ability to realize the full capability of the underlying model.

    \item \textbf{A new training paradigm for high-quality parallel generation.}
We introduce \emph{introspective-consistency training}, a simple yet effective recipe for converting pretrained AR models into introspective DLMs (e.g., \textbf{just using ~5B tokens}). By explicitly enforcing model's introspective consistency, our method enables parallel decoding without sacrificing AR-level quality. Unlike prior approaches, it requires neither distillation schedules nor masking curricula, yielding a stable and efficient path to high-quality DLMs.

\item \textbf{A novel single-pass decoding algorithm that unifies generation and verification.}
We propose \emph{Introspective Strided Decoding (ISD)}, which simultaneously generates new tokens and revises prior ones within the same forward pass. At \texttt{[MASK]} positions, the model proposes new tokens; at introspection positions, it revisits previous tokens against its causal anchor distribution. This yields outputs that provably match the base AR distribution, without confidence heuristics or separate verification passes.


\item \textbf{An AR-compatible serving stack for deployable DLMs and self-speculative decoding}
We design an inference stack that is directly compatible to existing AR serving systems (e.g., SGLang). 
Besides, we develop a gated residual adaptation mechanism where LoRA adapters are applied only at mask positions, while verification relies on the base model weights. It supports a continuum between near-lossless and strictly lossless modes.

\item \textbf{Comprehensive evidence that closes the quality gap while delivering real efficiency gains.}
Across 15 benchmarks, we show that I-DLM is the first DLM to match strong same-scale AR quality while substantially outperforming prior DLMs in both capability and serving efficiency. Our results establish a new quality--efficiency frontier.

\end{denseitemize}

We will open-source models and systems to facilitate community research and deployment.

%% file: sections/motivation.tex
\section{Background and Motivation}
\label{sec:motivation}

\begin{figure*}[t]
\begin{center}
\begin{subfigure}[t]{0.32\textwidth}
\centering
\includegraphics[width=\textwidth]{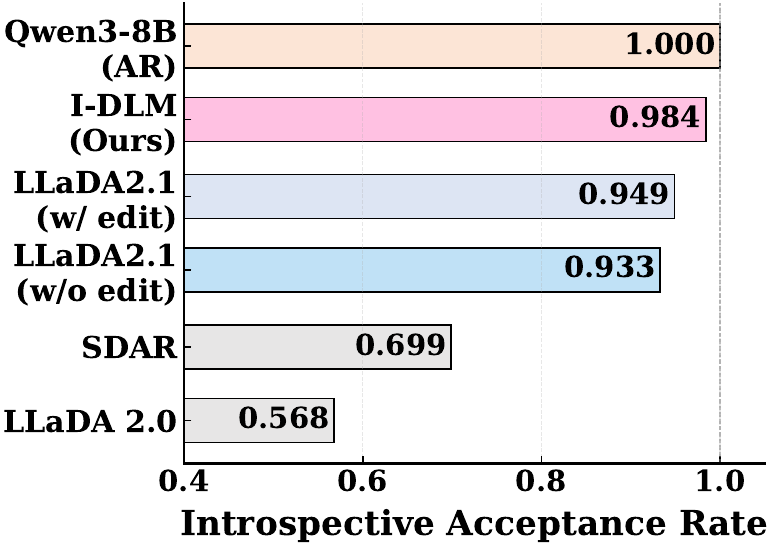}
\caption{\textbf{Generation--introspection gap.} Introspective acceptance rate (avg $\min(1, p/q)$) between generation ($q$) and introspection ($p$) distributions. AR and Ours achieve near-perfect consistency.}
\label{fig:introspection_rate}
\end{subfigure}
\hfill
\begin{subfigure}[t]{0.32\textwidth}
\centering
\includegraphics[width=\textwidth]{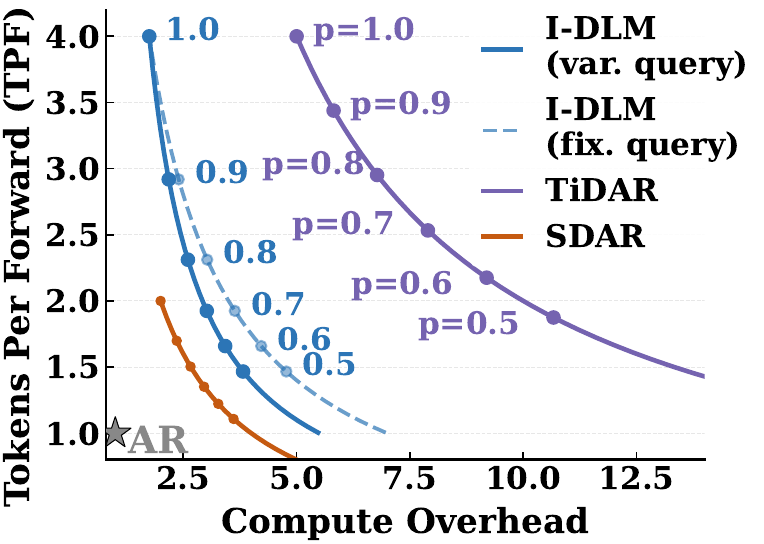}
\caption{\textbf{Compute economics} at stride $N{=}4$. $p$ denotes the per-token acceptance probability. ISD achieves high TPF at low overhead. 
}
\label{fig:compute_bottleneck}
\end{subfigure}
\hfill
\begin{subfigure}[t]{0.32\textwidth}
\centering
\includegraphics[width=\textwidth]{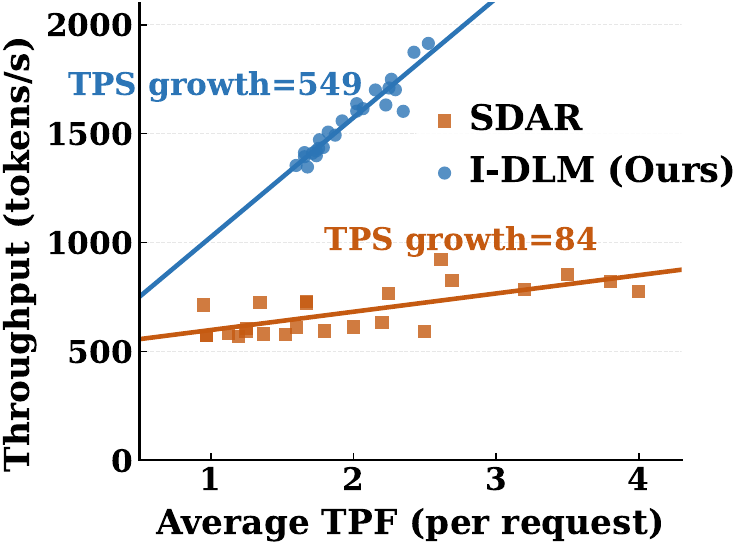}
\caption{\textbf{Batching efficiency.} TPF vs.\ batch throughput (tokens/s) at batch size 8. SDAR's throughput barely scales with TPF; I-DLM translates higher TPF into proportionally higher throughput.}
\label{fig:infra_bottleneck}
\end{subfigure}
\end{center}
\vspace{-.3cm}
\caption{\textbf{Bottleneck analysis.} Key gaps between DLMs and AR models: (a)~existing DLMs exhibit a generation--introspection gap---they can generate tokens but cannot reliably introspect on their own output, as measured by the introspection rate; (b)~DLM parallel decoding consumes far more compute per token, collapsing throughput under concurrency; (c)~higher TPF does not translate to proportionally higher throughput for existing DLMs.}
\label{fig:bottlenecks}
\vspace{-.3cm}
\end{figure*}

We trace these gaps to three inherent limitations of current DLMs (Figure~\ref{fig:bottlenecks}). (1) Low introspective consistency: DLMs cannot reliably agree with their generations, making coherent reasoning difficult; (2) Compute inefficiency: both training and inference require substantially more FLOPs in tokens per forward (TPF) than AR models; and (3) Inference engine incompatibility: multi-token, multi-step denoising in DLM inference is poorly aligned with modern AR serving stacks, leading to inefficient execution.

\textbf{DLMs lack introspective consistency.}
\label{sec:mot_quality}
We formalize \emph{introspective acceptance rate} $\alpha$ as follows: for each generated token $x_k$ sampled from distribution $q_k$, we perform a separate forward pass with all tokens revealed to obtain the corresponding causal distribution $p_k$ and compute $\alpha = \frac{1}{L}\sum_{k} \min(1,\, p_k(x_k)/q_k(x_k))$. For AR models, $p = q$ by construction, yielding $\alpha = 1$ and thus perfect generation-introspection consistency.
We evaluate $\alpha$ on IFEval using each model's
best configuration from its official release (Figure~\ref{fig:introspection_rate}).
SDAR (8B) achieves only 0.699 and LLaDA~2.0-flash (8B) only 0.568, indicating substantial divergence between what these models generate and what they would endorse upon re-examination. LLaDA~2.1~\citep{bie2026llada2} is philosophically close to us. Although it is motivated as improving the revision capability of DLMs, we view revision and introspective acceptance as deeply connected: improving a model's ability to revise its own outputs naturally improves its ability to endorse them upon re-examination. LLaDA~2.1 restructures the training and data pipeline for a token-to-token supervision format, whereas our introspective-consistency training is much simpler, jointly training generation and introspection under a unified objective without redesigning the data pipeline (\S\ref{sec:training}). Despite this simplicity, I-DLM achieves substantially higher introspective acceptance rates with far greater token efficiency: it matches its AR base model using only 4.5B training tokens, while SDAR needs 54B tokens ($12\times$ more) and still yields much worse downstream quality (10.0 vs.\ 69.6 on AIME-24).

\textbf{Parallel decoding does not translate to compute efficiency.}
\label{sec:mot_compute}
Existing baselines incur more flops per decoded token, which effectively dilutes the achieved speedup by pushing the kernel to a compute bound, as shown in Figure~\ref{fig:compute_bottleneck}. We define compute overhead as the ratio of total FLOPs to decode a given number of tokens compared to AR decoding. We provide detailed tokens per forward (TPF) pass and compute overhead analysis given different per-token acceptance rates in Appendix~\ref{app:tpf_derivation}. 
Figure~\ref{fig:compute_bottleneck} shows the tradeoff at stride $N{=}4$ for I-DLM, SDAR (block diffusion model), and Tidar (DLM with branched inference methods): at a TPF of ${\sim}2.5$, I-DLM incurs only ${\sim}2.5\times$ compute overhead, while TiDAR requires ${\sim}7.8\times$. In contrast, SDAR's TPF is capped at 2.0. This is because block diffusion (SDAR, LLaDA) requires $T$ denoising steps plus a mandatory KV-commit forward that produces no new tokens, capping TPF at $N/2$ even under ideal acceptance ($p{=}1$). Appendix~\ref{app:kv_commit} details why this overhead is hard to eliminate.

\begin{wrapfigure}{r}{0.32\columnwidth}
\vspace{-0.4cm}
\centering
\includegraphics[width=0.30\columnwidth]{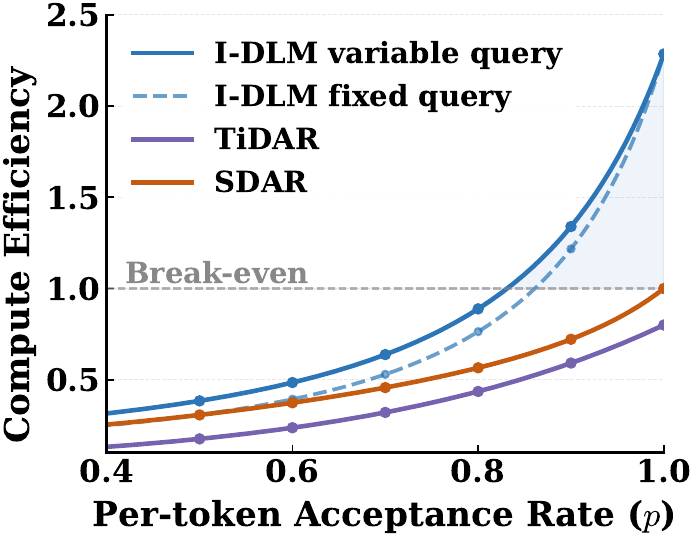}
\caption{\textbf{Compute efficiency} (TPF/OH) vs.\ acceptance rate at $N{=}4$. ISD is the only method above the break-even line.}
\label{fig:compute_efficiency}
\vspace{-0.4cm}
\end{wrapfigure}
To quantify this tradeoff, we define \emph{compute efficiency} as $\text{TPF} / \text{OH}$; efficiency $> 1$ means the TPF gain outweighs the overhead cost. Figure~\ref{fig:compute_efficiency} plots efficiency against acceptance rate $p$ for $N{=}4$. ISD is the only method that crosses the break-even line, reaching efficiency $> 1$ at $p \approx 0.83$ (variable-query) and $p \approx 0.86$ (fixed-query). At our empirically observed rates ($p \geq 0.85$), ISD achieves $1.08$--$2.29\times$ efficiency.
In contrast, TiDAR's efficiency is capped at $N/(N{+}1) = 0.80$ even at $p = 1$, and SDAR remains around $0.64$--$0.72$ at practical acceptance rates.

\textbf{DLM inference is incompatible with AR serving engines.}
The modern LLM serving stack (e.g., continuous batching, fused attention kernels, paged KV cache) is highly optimized for AR decoding. DLMs break these assumptions. In AR serving, continuous batching works because all requests advance uniformly. In block diffusion, tokens within a block converge at different rates: some positions pass the confidence threshold early, yet all requests must synchronize at the slowest block, wasting the TPF gain. Figure~\ref{fig:infra_bottleneck} reflects this: SDAR's TPS growth rate (slope$=$84) is nearly flat with respect to TPF. In contrast, ISD uses strict causal attention (compatible with AR kernels) and adaptive stride, where each step produces at least one quality-guaranteed token via introspection, achieving a TPS growth rate of 549 (\S\ref{sec:infra}). We show other AR serving mismatch (e.g., attention kernel mismatch) in Appendix~\ref{app:attn_kernel}. 

We bring up more discussion on detailed related works in Appendix~\ref{app:related}.


%% file: sections/method.tex
\section{Introspective Diffusion Language Model}
\label{sec:method}
We present I-DLM to address the aforementioned bottlenecks. Causal attention with logit shift closes the generation–introspection gap and improves training efficiency (\S\ref{sec:training}); introspective strided decoding eliminates the compute overhead of iterative denoising (\S\ref{sec:inference}); and the preserved causal structure enables direct integration into AR serving stacks (\S\ref{sec:infra}).

\subsection{Introspective-Consistency Training}
\label{sec:training}

We convert a pretrained AR model into an introspective diffusion model by combining causal attention, logit shift, and an all-masked objective. The training mask is shown in Appendix~\ref{app:attn_mask}. We note that our proposed introspective-consistency training method itself is simple, but it is built on a deep and fundamental insight inspired by AR:  AR training unifies generation and introspection in one forward pass; DLM training should inherit the same spirit.


\textbf{Causal training with logit shift.}
We apply \emph{strict causal masking} uniformly across both the masked and clean portions: for any query position $j$ and key position $i$, attention is permitted only when $i \leq j$. Unlike SDAR which uses block-causal attention (bidirectional within blocks), this holds for masked tokens in $x_t$ as well---each \texttt{[MASK]} attends only to preceding positions. In the clean region $x_0$, standard causal attention is used to ensure generation and introspection operate under the same attention pattern (Section~\ref{sec:mot_quality}) and enable KV cache reuse at inference. We pair this with a logit shift: the hidden state at position $i$ is trained to predict token $i{+}1$ rather than token $i$. Standard masked diffusion trains $\text{hidden}[\texttt{M}_i] \to \text{token}[i]$, which breaks the AR model's inherent $\text{logits}[i] \to \text{token}[i{+}1]$ mapping and crucially prevents clean positions from providing a meaningful verify signal. Our shifted formulation preserves this mapping: clean positions produce the causal anchor $p$ (the verify distribution), while masked positions produce tokens $q$ (the decode distribution).

\textbf{All-masked training with auto-balanced loss.}
Given a clean sequence $x_0 = (x_1, \ldots, x_L)$, we replace \emph{all} tokens with a special \texttt{[MASK]} token to obtain the fully masked input $x_t = (\texttt{M}, \ldots, \texttt{M})$. The training input is the concatenation $[x_t \,|\, x_0]$, where $x_t$ is the all-masked sequence and $x_0$ provides the clean reference. Unlike standard masked diffusion training, which masks a random fraction $r$ of tokens, wasting $(1{-}r)$ of the compute on unsupervised positions, our all-masked regime ensures that every position contributes a useful training signal, eliminating the supervision dilution identified in Section~\ref{sec:mot_quality}.

We apply a cross-entropy loss with shifted labels separately to the masked and clean regions:
\begin{equation}
    \mathcal{L}_{\text{mask}} = -\frac{1}{|\mathcal{S}_t|}\sum_{\ell \in \mathcal{S}_t} \log p_\theta(x_0^{\ell+1} \mid [x_t, x_0]_{\leq \ell}), \quad
    \mathcal{L}_{\text{clean}} = -\frac{1}{|\mathcal{S}_0|}\sum_{\ell \in \mathcal{S}_0} \log p_\theta(x_0^{\ell+1} \mid [x_t, x_0]_{\leq \ell}),
    \label{eq:loss_split}
\end{equation}
where $\mathcal{S}_t$ and $\mathcal{S}_0$ are the non-padding positions in the masked and clean regions, respectively, and $x_0^{\ell+1}$ is the shifted target (next token). This is the same cross-entropy objective used in AR pretraining---no separate distillation loss or teacher model is required. On clean positions, $\mathcal{L}_{\text{clean}}$ trains the \emph{introspection} pathway: the model learns to produce the causal anchor distribution $p_\theta(x_{i+1} \mid x_{\leq i})$, recovering the exact AR training objective. On masked positions, $\mathcal{L}_{\text{mask}}$ trains the \emph{decode} pathway: the model learns to produce tokens $q$ from \texttt{[MASK]} hidden states, enabling strided generation at stride $N > 1$.

Although both pathways share the same cross-entropy objective, their loss magnitudes can differ substantially: masked positions face a harder prediction task and tend to produce larger losses, especially early in training. A fixed weighting risks the decode pathway dominating the gradient, undermining the introspection pathway that is critical for verification quality. We address this with an auto-balanced loss:
\begin{equation}
    \mathcal{L} = \mathcal{L}_{\text{mask}} + \hat{s} \cdot \mathcal{L}_{\text{clean}}, \quad \hat{s} = \frac{\mathcal{L}_{\text{mask}}}{\mathcal{L}_{\text{clean}}},
    \label{eq:loss}
\end{equation}
where $\hat{s}$ is the ratio of loss magnitudes, treated as a fixed scalar at each training step (not differentiated through). This dynamically rescales the clean-position loss to match the magnitude of the masked-position loss, ensuring both pathways receive equal effective gradient magnitude without manual tuning. 
Because both pathways share the same objective and receive balanced supervision, the model naturally aligns $q$ with $p$, maximizing the introspective acceptance rate (Figure~\ref{fig:introspection_rate}).

\noindent\textbf{A note on why previous works fail to achieve introspective consistency:} 
While the individual ingredients of our approach, such as causal-mask training \citep{hu2025fast}, logit shifting \citep{ye2025dream}, and full-mask training \citep{liu2025tidar}, have each been explored in isolation, their combined role in enabling true introspective consistency has remained largely overlooked. We ablate these components in Figure~\ref{fig:training_ablation}: removing causal attention and logit shift (reverting to block diffusion) causes sharp degradation on reasoning tasks (e.g., HumanEval: 92.7 $\rightarrow$ 60.3), confirming that introspective consistency is critical for long-horizon generation. Specifically, causal masking ensures generation-time context consistency across denoising steps, logit shifting further bridges the verification and generation with unified hidden states and ensures training robustness by respecting the autoregressive (AR) model's inherent behavior, and the all-masked objective ensures training efficiency through dense supervision; no single component achieves consistency, robustness, and efficiency simultaneously. 
Prior work such as NBDiff \citep{tian2025next} adopts causal masking for prefilling and block-causal masking for decoding, and achieves substantial gains. This confirms that even partial generation-time consistency at the prefilling stage alone delivers notable improvements, but switching between masking schemes introduces nontrivial computational overhead at inference time. Similarly, LLaDA-2.1 shares our broader goal of revision during generation, but it depends on heavy data engineering to construct multi-turn revision sequences. In contrast, our method provides a unified training pathway that ensures consistency, robustness, and efficiency simultaneously by directly distilling the AR model’s inherent behavior through an efficient single-stage training regime.



\subsection{Introspective Strided Decoding (ISD)}
\label{sec:inference}

\begin{figure*}[t]
\begin{center}
\includegraphics[width=0.95\textwidth]{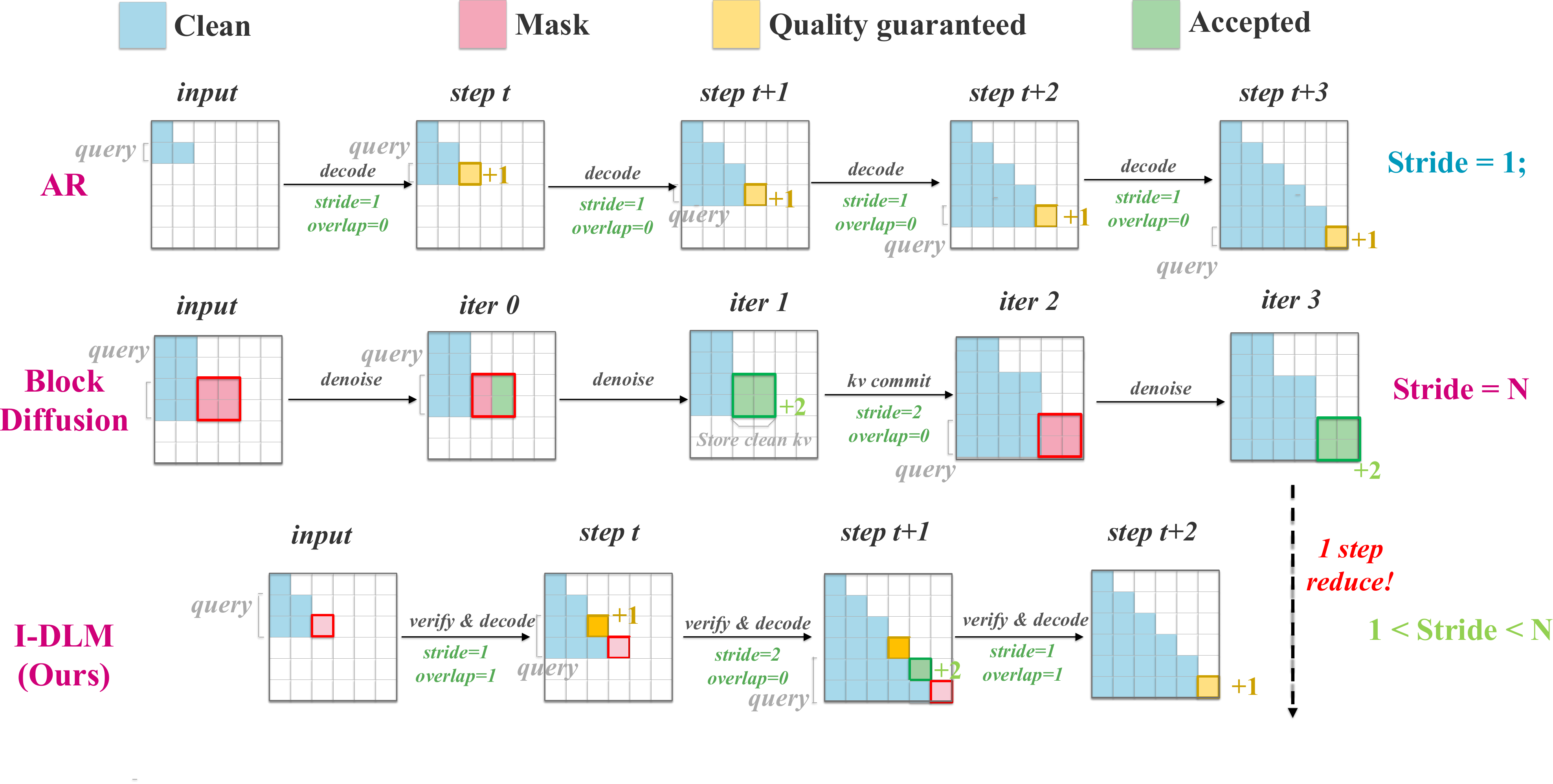}
\end{center}
\vspace{-.7cm}
\caption{\textbf{Comparison of decoding paradigms.}
Our I-DLM uses strict causal attention with adaptive stride ($1 < \text{stride} < N$) and is a drop-in replacement within AR serving infrastructure. ISD produces a quality-guaranteed token $\bm{x}_{i+1}$ together with draft tokens $\hat{\bm{x}}_{i+2:i+N}$ via introspective strided decoding, $\bm{x}_{i+1}, \hat{\bm{x}}_{i+2:i+N} = \pi_{\text{strided}}(c_{1:i}, m_{1:N-1})$; Residual ISD (R-ISD) additionally gates a LoRA residual for bit-for-bit lossless output: $\bm{x}_{i+1:i+N} = \pi_{\text{ar}}(c_{1:i}, m_{1:N-1}) + \pi_{\text{L}}(m_{1:N-1})$.
 }
\label{fig:inference_block_n}
\vspace{-0.6cm}
\end{figure*}


Introspective Strided Decoding (ISD) is the inference counterpart to I-DLM training. Because the model uses strict causal attention, ISD begins with a standard AR prefill over the prompt, then \emph{dynamically} selects the effective stride at each decode step via the $p/q$ acceptance criterion. Unlike block diffusion which commits to a fixed block size, or speculative decoding which requires a separate draft model, the stride in ISD adapts intrinsically based on the model's own self-consistency: easy tokens are accepted in parallel while difficult tokens fall back toward AR-quality generation.

\textbf{Single-pass stride and introspection.}
ISD generates tokens in steps (Algorithm~\ref{alg:spec_verify}, Figure~\ref{fig:inference_block_n} bottom): 
\blackcircled{1} \emph{Step 1 (bootstrap).}
We append $N$ \texttt{[MASK]} tokens to the prefix and run a single forward pass. Due to the logit shift, the last clean position produces a quality-guaranteed token $x_1$---identical to an AR prediction, requiring no introspection. This token anchors the subsequent introspection chain: at the next step, $x_1$'s hidden state produces the causal anchor $p_2$ for evaluating $\hat{x}_2$, and so on. The remaining $N{-}1$ \texttt{[MASK]} positions produce strided tokens $\hat{x}_2, \ldots, \hat{x}_N$; and \blackcircled{2}
\emph{Step $t > 1$ (stride + introspection).}
We fill in the accepted (or resampled) tokens from the previous step, append $N$ fresh \texttt{[MASK]} tokens, and run a \emph{single} forward pass that simultaneously:
\begin{denseitemize}
    \item \emph{Introspects} on the previous tokens, the filled-in tokens are now clean, so the logits at each position produce causal anchor distributions $p_k$ (the true AR next-token distributions);
    \item \emph{Generates} new tokens from the appended \texttt{[MASK]} positions.
\end{denseitemize}
Crucially, introspection piggybacks on the next stride step at zero additional cost: every step after the first produces both accepted tokens and fresh tokens in one forward pass. This contrasts with SDAR ($T{+}1$ forwards per block), WeDLM (confidence-based streaming without distributional guarantees), and TiDAR (unified draft and introspection but $N(N{+}1)$ token queries per step; our logit shift reduces this to $N^2$).

\textbf{Adaptive stride via $p/q$ acceptance.}
We introspect on each token by comparing it against its causal anchor using the $p/q$ acceptance criterion~\citep{leviathan2023fast}: token $x_k$ sampled from $q_k$ is accepted with probability $\min(1, \, p_k(x_k) / q_k(x_k))$.
When accepted, the token provably follows the causal anchor distribution.
On rejection, the token is resampled from the corrected distribution $\text{normalize}(\max(0, p_k - q_k))$ and all subsequent tokens are discarded; the next step becomes a pure stride with no tokens to introspect.
When all tokens are accepted, a bonus token is sampled from the final anchor distribution, achieving stride $N{+}1$.
This mechanism naturally adapts the effective stride per step without manual tuning. The formal procedure is given in Algorithm~\ref{alg:spec_verify}, with a detailed step-by-step illustration in Appendix~\ref{app:isd_steps}.

\begin{algorithm}[t]
\small
\caption{Introspective Strided Decoding (one iteration)}
\label{alg:spec_verify}
\begin{algorithmic}[1]
\REQUIRE Prefix tokens $x_{1:L}$, stride $N$, model $M$ with logit shift
\ENSURE Accepted new tokens appended to prefix
\STATE \textcolor{themecolor}{// Stride: $N{-}1$ masks produce $N$ tokens (logit shift gives +1 free)}
\STATE Construct input $[x_{1:L},\, \texttt{MASK}_1,\, \ldots,\, \texttt{MASK}_{N-1}]$ \hfill ($N{-}1$ masks)
\STATE Run $M$ $\rightarrow$ $\text{logits}_{\text{stride}}[1{:}L{+}N{-}1]$
\STATE Sample $x_{L+1} \sim \text{softmax}(\text{logits}_{\text{stride}}[L])$ \hfill \COMMENT{quality-guaranteed (AR prediction)}
\FOR{$k = 2$ \TO $N$}
    \STATE Sample $x_{L+k} \sim q_k \triangleq \text{softmax}(\text{logits}_{\text{stride}}[L{+}k{-}1])$ \hfill \COMMENT{strided proposal}
\ENDFOR
\STATE \textcolor{themecolor}{// Introspect: verify proposals against causal anchors}
\STATE \textcolor{themecolor}{// (Only introspection shown; in practice fused with next proposal into one $2N{-}1$ token pass)}
\STATE Construct input $[x_{1:L},\, x_{L+1},\, \ldots,\, x_{L+N}]$
\STATE Run $M$ $\rightarrow$ $\text{logits}_{\text{anchor}}[1{:}L{+}N]$
\STATE $p_k \leftarrow \text{softmax}(\text{logits}_{\text{anchor}}[L{+}k{-}1])$ for $k = 1, \ldots, N$ \hfill \COMMENT{causal anchor distributions}
\STATE \textcolor{themecolor}{// Introspection with adaptive stride}
\STATE $n_{\text{accepted}} \leftarrow 1$ \hfill \COMMENT{$x_{L+1}$ always accepted (quality-guaranteed)}
\FOR{$k = 2$ \TO $N$}
    \STATE Draw $r \sim \text{Uniform}(0, 1)$
    \IF{$r < \min\!\big(1,\; p_k(x_{L+k}) \,/\, q_k(x_{L+k})\big)$}
        \STATE Accept $x_{L+k}$;\quad $n_{\text{accepted}} \leftarrow n_{\text{accepted}} + 1$
    \ELSE
        \STATE Resample $x_{L+k} \sim \text{normalize}\!\big(\max(0,\; p_k - q_k)\big)$
        \STATE $n_{\text{accepted}} \leftarrow n_{\text{accepted}} + 1$;\quad \textbf{break}
    \ENDIF
\ENDFOR
\STATE \textcolor{themecolor}{// Bonus token if all proposals accepted}
\IF{$n_{\text{accepted}} = N$}
    \STATE Sample $x_{L+N+1} \sim \text{softmax}(\text{logits}_{\text{anchor}}[L{+}N])$
    \STATE $n_{\text{accepted}} \leftarrow n_{\text{accepted}} + 1$
\ENDIF
\STATE
\STATE Append $x_{L+1:L+n_{\text{accepted}}}$ to prefix; commit KV cache
\end{algorithmic}
\end{algorithm}

\textbf{Lossless ISD with residual adaptation.}
ISD uses a single model as both proposer and introspector, eliminating the need for a separate draft model to train, maintain, or synchronize. While it can be used as a standalone generative model, its single-forward drafting-and-verification structure also makes it a natural fit for self-speculative decoding.


Inspired by the gated LoRA approach in multi-token prediction~\citep{samragh2025your}, we gate the LoRA residual with a per-token binary mask: \texttt{[MASK]} positions use base+LoRA weights to produce high-quality strided tokens, while introspection positions use \emph{base-model-only} weights. Critically, because the entire model uses strict causal attention, introspection positions cannot attend to token positions---the causal anchor distribution $p$ is computed from base-only weights over a base-only KV cache, identical to a pure base AR forward pass. This guarantees that ISD introspects against the \emph{exact} base AR distribution, making the output bit-for-bit lossless (Appendix~\ref{app:gated_lora}).


\textbf{Theoretical speedup analysis.}
Let $p_k$ denote the acceptance probability of the $k$-th strided token, and $P_k = \prod_{j=1}^{k} p_j$ the cumulative acceptance probability. For ISD at stride $N$, the expected tokens per forward pass (TPF) is (see Appendix~\ref{app:tpf_derivation} for derivation):
\begin{equation}
\small
    \text{TPF}_N = \frac{2 + P_1 + P_2 + \cdots + P_{N-2}}{2 - P_{N-1}}.
    \label{eq:tpf}
\end{equation}
At perfect acceptance ($p_k = 1$), $\text{TPF}_N = N$ recovers the theoretical maximum. At $p_k = 0$, $\text{TPF}_N = 1$ degenerates to AR. In the memory-bound decode regime, forward pass latency is approximately constant regardless of stride size, so wall-clock speedup $\approx$ TPF. With typical acceptance rates of $p \geq 0.85$, stride $N{=}3$ achieves $\text{TPF} \approx 2.3\text{--}2.4\times$ with a compute overhead of only ${\sim}2\times$—meaning ISD translates most of the parallel decoding benefit into real throughput gain with minimal wasted compute. Our evaluations show that I-DLM achieves close-to-optimal speedup empirically (Section~\ref{sec:experiments}).

\subsection{I-DLM Serving Stack: AR-Compatible Serving}
\label{sec:infra}

Prior DLMs rely on custom inference pipelines that forgo the optimizations accumulated in AR serving systems (\S\ref{sec:motivation}). Because I-DLM preserves strict causal attention, we integrate it directly into SGLang as a drop-in extension, inheriting paged KV cache, continuous batching, and tensor parallelism without modification. We illustrate the key system optimizations below (ablated in \S\ref{sec:ablation}).

\paragraph{(1) AR-inherited optimizations.}
Each ISD step maps to SGLang's native \emph{extend} mode, appending $2N{-}1$ tokens with causal attention. Unlike block diffusion---where requests within a batch must synchronize at the slowest converging block, breaking continuous batching (\S\ref{sec:motivation})---ISD produces at least one quality-guaranteed token every step, so all requests advance uniformly and continuous batching works unmodified. This is why I-DLM's throughput scales proportionally with concurrency while block diffusion methods plateau (Figure~\ref{fig:infra_bottleneck}). The causal structure further enables two key reuses. First, we capture the entire extend forward into a single CUDA graph, replaying it each step with only \texttt{input\_ids} and attention metadata updated in-place. Second, since our extended sizes are small (${\leq}9$ tokens), we replace the three-kernel attention cascade with a single paged-only kernel per layer, eliminating $2L$ redundant launches (Appendix~\ref{app:attn_kernel}).

\paragraph{(2) Stationary-batch scheduling.}
ISD has a strict dependency chain: $\text{forward} \!\to\! \text{verify} \!\to\! \text{trim} \!\to\! \text{prepare} \!\to\! \text{forward}$. This prevents the CPU--GPU overlap used in AR serving, making CPU overhead directly additive to step latency. We mitigate this with a stationary-batch decode loop that reuses the batch object across consecutive ISD steps, bypassing the full scheduler rebuild. Within this loop, KV slots are allocated via a single batched scatter, constant metadata is cached, and the ISD-specific KV trim-and-commit cycle frees rejected and MASK positions after each verification. Non-critical I/O is deferred to an overlap window during the next GPU forward.

\paragraph{(3) Kernel fusion and proposal optimization.}
The verification step is fused into a single Triton kernel with online softmax and Gumbel-max correction; the common accept path (${\sim}$78\% of positions) returns after one streaming pass, skipping correction entirely. Since ISD's $p/q$ criterion guarantees output correctness regardless of proposal quality, we use argmax for proposals to maximize acceptance rate without affecting output diversity. For lossless R-ISD (Appendix~\ref{app:gated_lora}), we implement segment-gated LoRA where a per-token binary mask gates the LoRA residual within the CUDA graph, with cuBLAS replacing segmented GEMV at small token counts and a dedicated CUDA stream overlapping the LoRA shrink with base projections.

%% file: sections/experiments.tex
\section{Experiments}
\label{sec:experiments}

\subsection{Evaluation Methodology}
We train two I-DLM variants: \textbf{I-DLM-8B} and \textbf{I-DLM-32B}, converted from Qwen3-8B and Qwen3-32B~\citep{yang2025qwen3}, respectively, using the all-masked causal training recipe (Section~\ref{sec:training}). Training uses \textbf{4.5B tokens} on \textbf{8 H100 GPUs}. For lossless ISD, we additionally train LoRA adapters (\textbf{rank 128}) on the same data. Full training details are in Appendix~\ref{app:training_details}.

\textbf{Baselines.}
We compare against two classes of state-of-the-art methods. \emph{(i) Diffusion LLMs:} LLaDA-2.1-mini (16B)\citep{bie2026llada2}, LLaDA-2.0-flash (100B)\citep{bie2025llada2}, LLaDA-2.1-flash (100B)\citep{bie2026llada2}, SDAR (8B, 30B-A3B)\citep{cheng2025sdar}, NBDiff (7B)\citep{tian2025next}, DREAM (7B)\cite{ye2025dream}, WeDLM (8B)\cite{liu2025wedlm}, LightningRL (8B), TiDAR (8B), Jacobi Forcing (7B)\citep{hu2026lightningrl}, Fast-dLLM (7B)\citep{wu2025fast}, Mercury Coder Small\citep{labs2025mercury}, and Gemini Diffusion; and \emph{(ii) Speculative decoding:} EAGLE-3~\citep{li2025eagle}.
We use our I-DLM AR counterparts, Qwen3-8B and Qwen3-32B, as baselines.

\textbf{Benchmarks.}
We evaluate on 15 benchmarks covering four domains: (i) \emph{Knowledge and Reasoning}: ARC-C\citep{clark2018think}, MMLU\citep{hendrycks2020measuring}, MMLU-Pro\citep{wang2024mmlu}, GPQA-D\citep{rein2024gpqa}, GPQA\citep{rein2024gpqa}; (ii) \emph{Math Reasoning}: GSM8K\citep{cobbe2021training}, MATH-500\citep{hendrycks2021measuring}, MathBench\citep{liu2024mathbench}, AIME-24\citep{aime}, AIME-25\citep{aime}; (iii) \emph{Code Generation}: HumanEval\citep{chen2021evaluating}, MBPP\citep{odena2021program}, LiveCodeBench-v6\citep{jain2024livecodebench}; and (iv) \emph{Instruction Following}: IFEval\citep{zhou2023instruction}. All with thinking mode enabled.

\textbf{Metrics.}
For quality, we report accuracy on each benchmark with average values over three runs, where baseline results are taken from their original papers where available. For efficiency, we report \emph{request-level tokens per second} (latency) and \emph{server-level tokens per second} (throughput) under varying concurrency levels. Details in Appendix~\ref{app:eval_details}.

\subsection{End-to-End Performance}
\label{sec:results}

\begin{table}[t]
\centering
\caption{\textbf{End-to-end quality.} Accuracy (\%) on 15 benchmarks. I-DLM results use ISD ($N{=}4$, sampling). \underline{Underline}: best non-AR result under 30B. $^\dagger$: best non-AR result under 100B.
}
\vspace{-0.3cm}
\label{tab:main_results}
\scriptsize
\setlength{\tabcolsep}{3pt}
\begin{tabular}{lccccc|cc|cc}
\toprule
& \textbf{LLaDA-2.1} & \textbf{LLaDA-2.0} & \textbf{LLaDA-2.1} & \textbf{SDAR} & \textbf{SDAR} & \textbf{I-DLM} & \textbf{Qwen3} & \textbf{I-DLM} & \textbf{Qwen3} \\
& \textbf{-mini} & \textbf{-flash} & \textbf{-flash} & \textbf{8B} & \textbf{30B-A3B} & \textbf{8B} & \textbf{8B} & \textbf{32B} & \textbf{32B} \\
\textit{Params} & \textit{16B} & \textit{100B} & \textit{100B} & \textit{8B} & \textit{30B} & \textit{8B} & \textit{8B} & \textit{32B} & \textit{32B} \\
\midrule
\multicolumn{10}{l}{\textit{Knowledge \& Reasoning}} \\
ARC-C & 90.2 & --- & --- & 91.9 & 93.2 & \underline{95.8} & 95.8 & 96.8$^\dagger$ & 97.2 \\
MMLU & 74.5 & --- & --- & 78.6 & 82.8 & \underline{82.4} & 83.5 & 86.8$^\dagger$ & 87.2 \\
MMLU-Pro & 64.8 & 74.8 & 76.6 & 56.9 & 61.5 & \underline{73.1} & 75.1 & 79.7$^\dagger$ & 80.1 \\
GPQA-D & 46.0 & --- & --- & 40.2 & 36.7 & \underline{55.6} & 58.9 & 62.1$^\dagger$ & 64.1 \\
GPQA & 53.3 & 62.3 & 67.3$^\dagger$ & --- & --- & \underline{54.9} & 55.4 & 58.7 & 65.0 \\
\midrule
\multicolumn{10}{l}{\textit{Math}} \\
GSM8K & 89.0 & --- & --- & 91.7 & 91.4 & \underline{95.0}$^\dagger$ & 96.0 & 94.9 & 94.7 \\
MATH-500 & 85.0 & --- & --- & 78.6 & 77.8 & \underline{96.8} & 95.8 & 97.6$^\dagger$ & 97.8 \\
MathBench & 84.2 & --- & --- & 76.9 & 79.3 & \underline{89.1} & 93.1 & 95.6$^\dagger$ & 95.5 \\
AIME-24 & 43.3 & --- & --- & 10.0 & 16.7 & \underline{69.6} & 73.1 & 83.3$^\dagger$ & 76.7 \\
AIME-25 & 43.3 & 60.0 & 63.3 & 10.0 & 10.8 & \underline{60.8} & 65.4 & 80.0$^\dagger$ & 80.0 \\
\midrule
\multicolumn{10}{l}{\textit{Code}} \\
HumanEval & 86.0 & --- & --- & 78.7 & 87.2 & \underline{93.3} & 95.1 & 96.3$^\dagger$ & 96.3 \\
MBPP & 82.1 & --- & --- & 72.0 & 71.6 & \underline{92.2} & 93.4 & 94.6$^\dagger$ & 95.7 \\
LCB-v6 & 30.4 & 42.5 & 45.4 & 16.6 & 21.7 & \underline{45.7} & 50.3 & 57.1$^\dagger$ & 58.3 \\
\midrule
\multicolumn{10}{l}{\textit{Instruction Following}} \\
IFEval & 83.2 & 82.6 & 83.6 & 61.4 & 60.6 & \underline{84.7} & 84.7 & 84.7$^\dagger$ & 84.5 \\
\bottomrule
\end{tabular}
\end{table}

Table~\ref{tab:main_results} summarizes I-DLM's quality; ablations on different stride settings are in \S\ref{sec:ablation}. 

\textbf{I-DLM surpasses the quality of strongest DLMs with comparable sizes.}
I-DLM consistently outperforms all existing DLMs, often by large margins. At 8B scale, I-DLM-8B exceeds LLaDA-2.1-mini (16B) across all benchmarks despite using half the parameters, with particularly large gains on reasoning and code tasks (e.g., +26.3 on AIME-24 and +15.3 on LiveCodeBench-v6). Compared to SDAR, which is built on the same Qwen3-8B base, I-DLM improves dramatically (e.g., 69.6 vs.\ 10.0 on AIME-24). 
At a larger scale, I-DLM-32B continues to outperform substantially larger models, surpassing LLaDA-2.1-flash (100B) by +16.7 on AIME-25 and +11.7 on LiveCodeBench-v6. 
Moreover, 
Table~\ref{tab:extended} shows that our I-DLM even outperforms proprietary DLMs on code generation, such as Mercury Coder Small (90.0 vs.\ 76.6) and Gemini Diffusion (89.6 vs.\ 76.0).


\textbf{I-DLM achieves comparable quality to AR models.}
Beyond outperforming prior DLMs, I-DLM is the first to match the quality of same-scale AR models. As shown in Table~\ref{tab:main_results}, I-DLM-8B achieves near-identical performance to Qwen3-8B, matching exactly on ARC-C, IFEval, remaining within $\sim$1 point on MMLU, even surpassing Qwen3-8b on Math-500. 
This result validates that introspective consistency is a key missing ingredient in prior DLMs: by aligning generation with verification through logit-shifted causal training, I-DLM recovers AR-level capability.




\begin{table}[t]
\centering
\caption{\textbf{Extended comparison} on benchmarks commonly reported across diffusion LLM methods. ``---'' indicates the result is not reported in the original paper.}
\vspace{-0.3cm}
\label{tab:extended}
\scriptsize
\begin{tabular}{lccccc}
\toprule
\textbf{Method} & \textbf{GSM8K} & \textbf{MMLU} & \textbf{HumanEval} & \textbf{MBPP} & \textbf{IFEval} \\
\midrule
Qwen3-8B (AR) & 96.0 & 83.5 & 95.1 & 93.4 & 84.7 \\
\midrule
NBDiff (7B) & 91.0 & 82.9 & 89.0 & 87.6 & 60.8 \\
Jacobi Forcing (7B) & 91.4 & --- & 83.5 & 70.4 & --- \\
WeDLM (8B) & 90.2 & 75.5 & 75.0 & 67.0 & --- \\
LightningRL (8B) & 90.3 & --- & 72.6 & 58.3 & --- \\
TiDAR (8B) & 80.4 & 76.6 & 57.9 & 65.4 & --- \\
DREAM (7B) & 81.0 & 70.6 & 57.9 & 58.8 & 62.5 \\
Fast-dLLM (7B) & 78.5 & --- & 43.3 & 28.2 & --- \\
Mercury Coder Small & --- & --- & 90.0 & 76.6 & --- \\
Gemini Diffusion & --- & --- & 89.6 & 76.0 & --- \\
\midrule
\textbf{Ours (8B)} & \textbf{95.0} & \textbf{82.4} & \textbf{93.3} & \textbf{92.2} & \textbf{84.7 }\\
\bottomrule
\end{tabular}
\vspace{-0.5cm}
\end{table}



\begin{figure*}[t]
\centering
\includegraphics[width=0.95\textwidth]{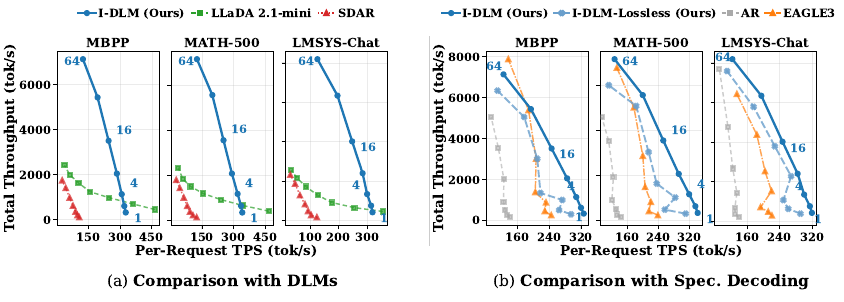}
\vspace{-.3cm}
\caption{\textbf{Throughput--latency tradeoff} across batch sizes (1, 4, 16, 64).
}
\label{fig:throughput}
\vspace{-.5cm}
\end{figure*}

\textbf{I-DLM delivers superior serving efficiency than existing DLMs.}
We evaluate end-to-end serving performance on MBPP, MATH-500, and LMSYS-Chat under concurrency levels $C \in \{1, 2, 4, 8, 16, 32, 64\}$ (Appendix~\ref{app:eval_details}). Figure~\ref{fig:throughput} shows the throughput–latency tradeoff. 
Across all workloads, I-DLM consistently outperforms prior DLMs at moderate to high concurrency. Starting from $C{\geq}4$, I-DLM achieves higher per-request throughput than both SDAR (8B) and LLaDA-2.1-mini (16B), with the advantage widening as concurrency increases. At typical deployment scales ($C{=}16$–32), I-DLM delivers 2.2–3.8$\times$ higher throughput than LLaDA-2.1-mini and 3.7–4.5$\times$ over SDAR. Under heavy load ($C{=}64$), I-DLM sustains stable per-request throughput ($\sim$125 tok/s), translating to 2.9–4.1$\times$ higher throughput at $C{=}64$.

\textbf{Comparison with speculative decoding for AR models.}
Figure~\ref{fig:throughput}(b) compares I-DLM against EAGLE3~\citep{li2025eagle}, a speculative decoding method that relies on an auxiliary draft model on top of the base AR model. I-DLM outperforms EAGLE3 in per-request throughput from $C{=}1$ through $C{=}32$ across all benchmarks (e.g., 341 vs.\ 238 tok/s on MATH-500, 319 vs.\ 221 on LMSYS-Chat, and 327 vs.\ 245 on MBPP at $C{=}1$). Notably, even I-DLM-Lossless---which produces output bit-for-bit identical to the base AR model---exceeds EAGLE3 at most concurrencies (e.g., 310 vs.\ 238 tok/s at $C{=}1$ on MATH-500), despite EAGLE3 requiring a separate draft model. As concurrency increases, I-DLM maintains its advantage over baselines (199 vs.\ 176 tok/s on MATH-500 and 195 vs.\ 184 on LMSYS-Chat at $C{=}32$).

\vspace{-0.3cm}
\subsection{Ablation Studies}
\label{sec:ablation}

\begin{figure*}[t]
\centering
\begin{subfigure}[t]{0.35\textwidth}
\centering
\includegraphics[width=\textwidth]{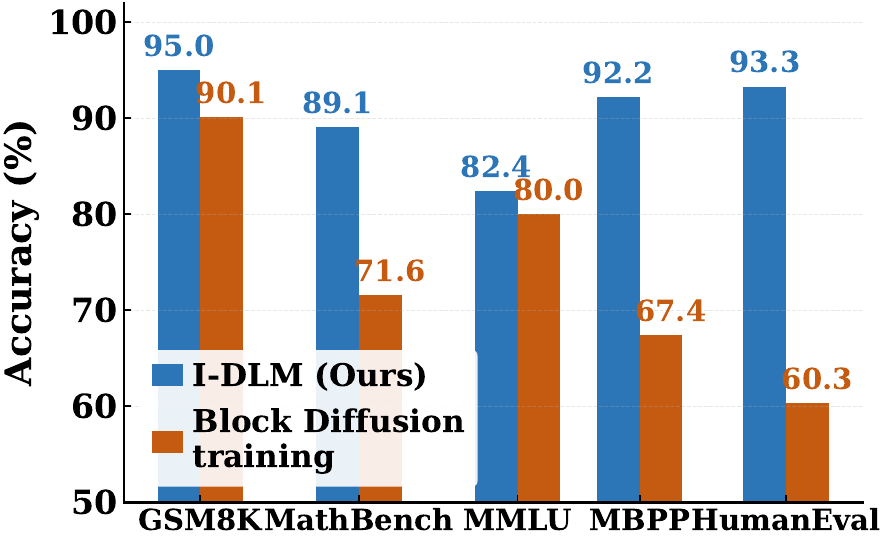}
\caption{\textbf{Training ablation.} I-DLM (causal + logit shift) vs.\ block diffusion (block-causal, no logit shift).}
\label{fig:training_ablation}
\end{subfigure}
\hfill
\begin{subfigure}[t]{0.62\textwidth}
\centering
\includegraphics[width=\textwidth]{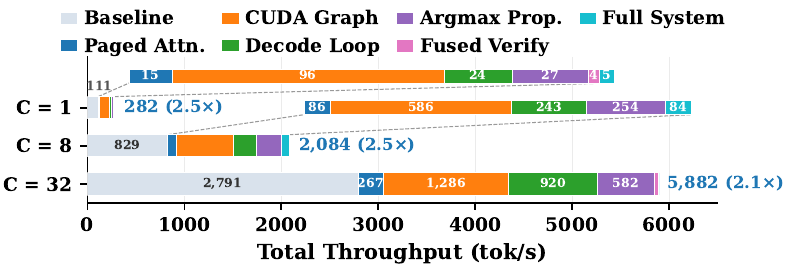}
\caption{\textbf{Systems optimization ablation.} Cumulative throughput at $C{=}1, 8, 32$. The full system achieves $2.1$--$2.5\times$ over the naive baseline.}
\label{fig:system_ablation}
\end{subfigure}
\vspace{-.1cm}
\caption{\textbf{Performance breakdown of the training design and systems optimizations.}}
\label{fig:ablation}
\end{figure*}

\vspace{-0.2cm}
\textbf{Ablations of the training design.}
Figure~\ref{fig:training_ablation} compares I-DLM's introspective-consistency training (causal attention + logit shift + all-masked objective) against standard block diffusion training (block-causal attention, no logit shift), on the same data budget. The gap is substantial, especially on long-horizon reasoning tasks: code generation drops sharply (HumanEval: 92.7 $\rightarrow$ 60.3; MBPP: 92.8 $\rightarrow$ 67.4), and math reasoning degrades significantly (MathBench: 89.1 $\rightarrow$ 71.6). Yet, knowledge tasks are relatively unaffected (MMLU: 82.4 $\rightarrow$ 80.0). This indicates that introspective consistency significantly reduces error accumulation over long reasoning chains.

\textbf{Ablation of the system design.}
Figure~\ref{fig:system_ablation} reports  efficiency breakdown at $C{=}1, 8, 32$. The largest gain comes from CUDA graph capture (+42–76\%), which eliminates kernel launch overhead. Decode-loop (stationary-batch scheduling) optimizations (+11–21\%) and argmax proposals (+11–15\%) further improve throughput by reducing host-side scheduling. 
Additional gains come from paged-only attention (+10–14\%) and kernel fusion (+1–4\%).


\begin{table}[h!]
\centering
\begin{minipage}{0.48\textwidth}
\vspace{-4pt}
\centering
\scriptsize
\caption{\textbf{Impact of stride size.} TPF, TPS (bs=1), and accuracy at stride $N$ on 1$\times$H100.}
\vspace{-0.3cm}
\label{tab:stride_scaling}
\begin{tabular}{ccccc}
\toprule
$N$ & TPF & TPS & MATH-500 & MBPP \\
\midrule
2 & 1.80 & 209.6 & 96.8 & 93.4 \\
3 & 2.48 & 281.9 & 95.8 & 92.8 \\
4 & 2.96 & 324.5 & 96.8 & 92.2 \\
8 & 4.01 & 445.1 & 94.6 & 88.3 \\
\bottomrule
\end{tabular}
\vspace{-8pt}
\end{minipage}
\hfill
\begin{minipage}{0.48\textwidth}
\vspace{-12pt}
\centering
\small
\caption{\textbf{Relaxed acceptance.} $\tau$ trades quality for TPF at $N{=}4$.}
\vspace{-0.3cm}
\label{tab:loose_threshold}
\begin{tabular}{lccccc}
\toprule
$\tau$ & 0 & 0.1 & 0.2 & 0.5 & 1\\
\midrule
HumanEval & 93.3 & 93.3 & 92.1 & 91.8 & 91.2 \\
TPF & 2.63 & 2.62 & 2.68 & 2.72 & 2.73 \\
\bottomrule
\end{tabular}
\vspace{-.2cm}
\end{minipage}
\end{table}

\textbf{Impact of stride size.}
The stride $N$ in ISD controls the parallelism-quality trade-off. Starting from our I-DLM-8B, we extend to larger strides ($N{=}8$) via continued training. As shown in Table~\ref{tab:stride_scaling}, TPF scales nearly linearly (1.80 $\rightarrow$ 4.01 from $N{=}2$ to $N{=}8$), while accuracy remains stable across tasks (e.g., MATH-500 within 94.6–96.8\%, MBPP within 88.3–93.4\%). 
These results show that I-DLM sustains quality under increased parallelism.


\textbf{Adaptive stride via relaxed acceptance.}
ISD offers a control knob: a loose threshold $\tau$ that multiplies the acceptance ratio by $(1{+}\tau)$, boosting the effective acceptance rate. At $\tau{=}0$ (strict), ISD provably matches the AR distribution. Increasing $\tau$ relaxes this guarantee for higher TPF without retraining.
As shown in Table~\ref{tab:loose_threshold}, quality is robust to relaxation: at $\tau{=}1.0$, HumanEval drops by only 1.6 points (93.3 $\rightarrow$ 91.2) while TPF increases from 2.63 to 2.73, suggesting that I-DLM's proposals are already well-aligned with the causal anchor.



%% file: sections/conclusion.tex
\vspace{-0.4cm}
\section{Conclusion}
\vspace{-0.3cm}
\label{sec:conclusion}

In this paper, we revisit the DLM design through the lens of autoregressive modeling and identify \emph{introspective consistency} as the missing principle behind their quality gap. Building on this insight, we introduce I-DLM, a new paradigm that unifies parallel generation and self-verification via logit-shifted causal training and introspective strided decoding. Our results show that enforcing consistency enables DLMs to match AR-level quality, substantially outperforming existing DLMs, achieving 2.9--4.1$\times$ better throughput on large concurrency.

%% file: sections/appendix.tex
\appendix

\section{Detailed Related work}
\label{app:related}

\paragraph{Diffusion Language Models.}

Masked diffusion language models corrupt text by replacing tokens with \texttt{[MASK]} and train a model to reverse the process~\citep{austin2021structured, sahoo2024simple, lou2023discrete}.
LLaDA~\citep{nie2025large} scaled this paradigm to 8B parameters, LLaDA~2.0~\citep{bie2025llada2} to 100B via mixture-of-experts, and LLaDA~2.1~\citep{bie2026llada2} introduced token editing with confidence-based decoding.
DREAM~\citep{ye2025dream} introduced the logit shift technique, aligning the diffusion objective with the AR model's $\text{logits}[i] \to \text{token}[i{+}1]$ mapping.
Block Diffusion~\citep{arriola2025block} generates fixed-size blocks autoregressively while denoising tokens within each block.

Converting pretrained AR models into diffusion models is a more data-efficient alternative.
\citet{gong2024scaling} showed that fine-tuning with a masked diffusion objective reduces training cost significantly.
Self-Distillation Through Time~\citep{deschenaux2024beyond} and Efficient-DLM~\citep{fu2025efficient} further streamline the pipeline.
SDAR~\citep{cheng2025sdar} converts AR models via full-model training on ${\sim}$50B tokens for block-parallel generation, and NBDiff~\citep{tian2025next} extends this with causal prefix constraints.
TiDAR~\citep{liu2025tidar} proposes a sequence-level hybrid that drafts via diffusion and verifies autoregressively.

Unlike these approaches, I-DLM combines strict causal attention with logit-shifted prediction from MASK positions---by maximally respecting the pretrained AR model's attention and prediction patterns, our conversion is far more data-efficient and closes the quality gap to the base AR model.


\paragraph{Speculative decoding and multi-token prediction for LLMs.}
Speculative decoding~\citep{leviathan2023fast, chen2023accelerating} accelerates AR models by drafting tokens with a fast model and verifying them in parallel, provably preserving the target distribution. This guarantee relies on the target model having a well-trained verify distribution $p$---a property that AR models possess inherently but standard DLLMs lack due to mask-only training.
Extensions include Medusa~\citep{cai2024medusa} (multiple prediction heads), the EAGLE family~\citep{li2024eagle, li2025eagle} (feature-level drafting), and SpecInfer~\citep{miao2023specinfer} (tree-structured verification).
Multi-token prediction (MTP) trains models to predict multiple future tokens simultaneously~\citep{gloeckle2024better}.
\citet{samragh2025your} proposed gated sparse expert LoRA for MTP---the closest prior work to our conditional LoRA, though designed for multi-head prediction rather than diffusion verification.
Consistency LLMs~\citep{kou2024cllms} achieve acceleration via Jacobi iteration.
ISD differs from these by using a single model with training-time stride capability to both achieve quality guarantee and deliver high latency and throughput gains under high concurrency.

\paragraph{DLLM-specific decoding.}
FastDLLM~\citep{wu2025fast} enables confidence-aware parallel decoding with KV cache reuse.
Jacobi Forcing~\citep{hu2025fast} distills diffusion models for fewer-step convergence but degrades at larger batch sizes.
Free Draft-and-Verification~\citep{wu2025free} explores self-speculative approaches.
WeDLM~\citep{liu2025wedlm} reconciles diffusion with causal attention for KV cache reuse.
These methods rely on confidence-based acceptance or iterative denoising, which either lacks formal quality guarantees or incurs high compute overhead; our ISD provides provable AR-quality output in a single stride--introspection cycle.

Despite these advances, fundamental gaps remain between DLLMs and AR models in training scalability, inference alignment, compute efficiency, and infrastructure compatibility. We analyze these gaps quantitatively in the next section.

\section{TPF and Compute Overhead Analysis}
\label{app:tpf_derivation}

We analyze the theoretical tokens per forward pass (TPF) and compute overhead for three parallel decoding paradigms: ISD (ours), block diffusion (SDAR), and branched self-speculative decoding (TiDAR). Throughout, $N$ is the block/stride size and $p$ is the uniform per-token acceptance probability ($P_k = p^k$). Compute overhead (OH) is the ratio of total query tokens to output tokens; AR has OH${}=1$ by definition.

\subsection{ISD (Ours)}

ISD alternates between an \emph{NP step} (propose only: append $N$ masks, produce 1 free token via logit shift + $N{-}1$ speculative proposals, finalize nothing) and \emph{P steps} (introspect previous proposals + propose new ones). A renewal cycle is one NP step followed by a geometric chain of P steps ending on the first rejection.

Each P step has probability $p^{N-1}$ of all-pass (finalizing $N$ tokens, continuing the chain) and $1 - p^{N-1}$ of rejection. On rejection at position $k$, we finalize $k + 1$ tokens (1 free + $k{-}1$ accepted + 1 resampled).

\paragraph{TPF.} Expected tokens and forwards per cycle:
\begin{equation}
    \mathbb{E}[\text{tokens}] = \frac{2 + p + \cdots + p^{N-2}}{1 - p^{N-1}}, \qquad
    \mathbb{E}[\text{forwards}] = \frac{2 - p^{N-1}}{1 - p^{N-1}}.
\end{equation}
\begin{equation}
    \text{TPF}_\text{ISD} = \frac{2 + p + p^2 + \cdots + p^{N-2}}{2 - p^{N-1}}.
\end{equation}
At $p = 1$: $\text{TPF} = N$; at $p = 0$: $\text{TPF} = 1$ (AR).

\paragraph{Overhead.} The P step processes $2N{-}1$ query tokens ($N{-}1$ filled proposals + $N$ masks). The NP step processes $N$ queries (variable) or $2N{-}1$ queries (fixed, padded to match P).

\emph{Variable query:}
\begin{equation}
    \text{OH}_\text{var} = \frac{3N - 1 - Np^{N-1}}{2 + p + \cdots + p^{N-2}}.
\end{equation}

\emph{Fixed query} (SGLang/vLLM, both steps use $2N{-}1$):
\begin{equation}
    \text{OH}_\text{fix} = \frac{(2N{-}1)(2 - p^{N-1})}{2 + p + \cdots + p^{N-2}}.
\end{equation}

The gap vanishes as $p \to 1$ and is irrelevant in the memory-bound regime where wall-clock speedup $\approx$ TPF.

\subsection{Block Diffusion (SDAR)}

SDAR generates a block of $N$ tokens via iterative denoising with a force schedule, then commits the KV cache in a separate forward pass. At each denoising step, $H \sim \text{Binomial}(R, p)$ tokens pass the confidence threshold out of $R$ remaining; at least $\max(H, 1)$ are committed. Let $\mathbb{E}[S \mid N]$ be the expected denoising steps for $N$ tokens (computed recursively). Total forwards = $\mathbb{E}[S \mid N] + 1$ (denoising + KV commit), each processing all $N$ tokens.

\begin{equation}
    \text{TPF}_\text{SDAR} = \frac{N}{\mathbb{E}[S \mid N] + 1}, \qquad
    \text{OH}_\text{SDAR} = \mathbb{E}[S \mid N] + 1.
\end{equation}

Note $\text{TPF} \times \text{OH} = N$ always. Even at $p = 1$ (1 denoising step + 1 KV commit = 2 forwards), $\text{TPF}_\text{SDAR} \leq N/2$---the mandatory KV commit caps throughput at half the theoretical maximum.

\subsection{Branched Self-Speculative Decoding (TiDAR)}

TiDAR uses a single forward to both verify $N$ draft tokens and pre-draft $N$ branches of $N$ masks each, covering all possible rejection outcomes. The input is $N$ verify tokens + $N^2$ branch masks = $N(N{+}1)$ queries per forward, always exactly 1 forward per cycle.

\begin{equation}
    \text{TPF}_\text{TiDAR} = 1 + p + p^2 + \cdots + p^{N-1} = \frac{1 - p^N}{1 - p}.
\end{equation}
\begin{equation}
    \text{OH}_\text{TiDAR} = \frac{N(N{+}1)}{1 + p + \cdots + p^{N-1}}.
\end{equation}

TiDAR achieves the highest TPF (no NP recovery step), but the $N^2$ branch masks are structurally wasteful: only 1 of $N$ branches is selected per forward, so $(N{-}1) \cdot N$ mask queries are always discarded. Even at $p = 1$, $\text{OH} = N{+}1$ and efficiency $= N/(N{+}1) < 1$---TiDAR can never be FLOP-efficient.

\subsection{Summary}

\begin{table}[h]
\centering
\small
\caption{TPF and overhead formulas (uniform $p$, stride $N$).}
\label{tab:tpf_summary}
\begin{tabular}{lccc}
\toprule
\textbf{Method} & \textbf{TPF} & \textbf{Overhead} & \textbf{Queries/fwd} \\
\midrule
ISD (var.) & $\frac{2+p+\cdots+p^{N-2}}{2-p^{N-1}}$ & $\frac{3N-1-Np^{N-1}}{2+p+\cdots+p^{N-2}}$ & $N$ or $2N{-}1$ \\[6pt]
ISD (fix.) & (same) & $\frac{(2N{-}1)(2-p^{N-1})}{2+p+\cdots+p^{N-2}}$ & $2N{-}1$ \\[6pt]
SDAR & $\frac{N}{\mathbb{E}[S|N]+1}$ & $\mathbb{E}[S|N]+1$ & $N$ \\[6pt]
TiDAR & $\frac{1-p^N}{1-p}$ & $\frac{N(N+1)(1-p)}{1-p^N}$ & $N(N{+}1)$ \\
\bottomrule
\end{tabular}
\end{table}

\section{Why Block Diffusion Requires a Separate KV Commit Pass}
\label{app:kv_commit}

Our SDAR inference uses SGLang's native DLLM support\footnote{\url{https://github.com/sgl-project/sglang/pull/19044}}. In this section we explain why block diffusion methods (SDAR, LLaDA) require a mandatory KV commit forward pass after denoising, and why this overhead is difficult to eliminate.

After $T$ denoising steps produce $N$ final tokens, a separate forward pass with \texttt{store\_kv=True} must write these tokens into the KV cache so that subsequent blocks can attend to them. This pass produces zero new tokens and is mandatory even at $p{=}1$, capping TPF at $N/2$.

A natural optimization would fuse commit with the next block's denoising in one forward: $[\underbrace{t_1, \ldots, t_N}_{\text{commit}},\, \underbrace{\texttt{M}, \ldots, \texttt{M}}_{\text{denoise}}]$. However, this requires \emph{per-position mixed attention masks}: committed tokens need bidirectional attention among themselves, while MASK tokens need block-causal attention. Current attention kernels (FlashAttention, FlashInfer) support only a single global \texttt{causal} flag and cannot mix bidirectional and causal attention within one forward pass. Implementing this would require custom attention kernels, doubled query size ($2N$), and complex batching logic for concurrent requests at different stages. SGLang's codebase confirms this: \texttt{dllm\_is\_commit} and \texttt{dllm\_needs\_commit} flags are defined but never activated---the optimization was considered but abandoned.

Because I-DLM uses strict causal attention throughout, no commit pass is needed. Each ISD step is a standard extend operation that incrementally commits KV entries as in AR decoding---no mixed masks, no custom kernels.

\section{Attention Kernel Overhead}
\label{app:attn_kernel}

Block diffusion uses block-causal attention, which differs from the strict token-level causal mask that AR serving kernels are optimized for. In SGLang, the standard extend path uses a cascade of three attention kernels per layer: (1) ragged attention among new tokens, (2) paged attention against the cached prefix, and (3) a merge kernel to combine the two via log-sum-exp renormalization. This cascade is optimized for large prefills but wasteful for DLLM decode steps that process only $N{=}4$--$5$ tokens, where the ragged kernel's advantage vanishes and the $3\times$ kernel launch overhead dominates.

\begin{figure}[h]
\centering
\includegraphics[width=0.5\columnwidth]{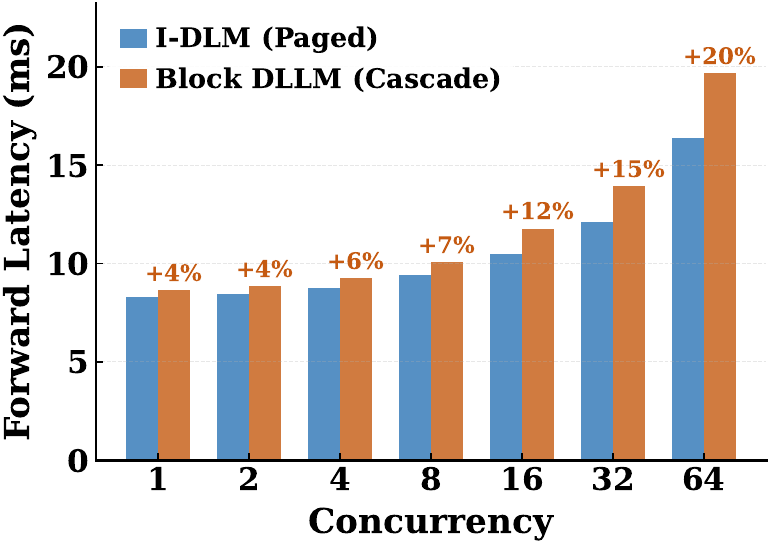}
\caption{\textbf{Attention kernel forward latency} at varying concurrency. I-DLM (Paged, single kernel) vs.\ Block DLLM (Cascade, three kernels). The cascade overhead grows from $+4\%$ at $C{=}1$ to $+20\%$ at $C{=}64$.}
\label{fig:attn_kernel}
\end{figure}

Because I-DLM uses strict causal attention with small extended sizes, it can bypass the cascade and use a single paged attention kernel per layer, reducing kernel launches from $3L$ to $L$ (where $L$ is the number of layers). Figure~\ref{fig:attn_kernel} shows the resulting forward latency comparison across concurrency levels. At low concurrency ($C{=}1$), the cascade overhead is modest ($+4\%$), but it grows to $+20\%$ at $C{=}64$ as the additional kernel launch overhead compounds with batching.

\section{Attention Mask Structure}
\label{app:attn_mask}

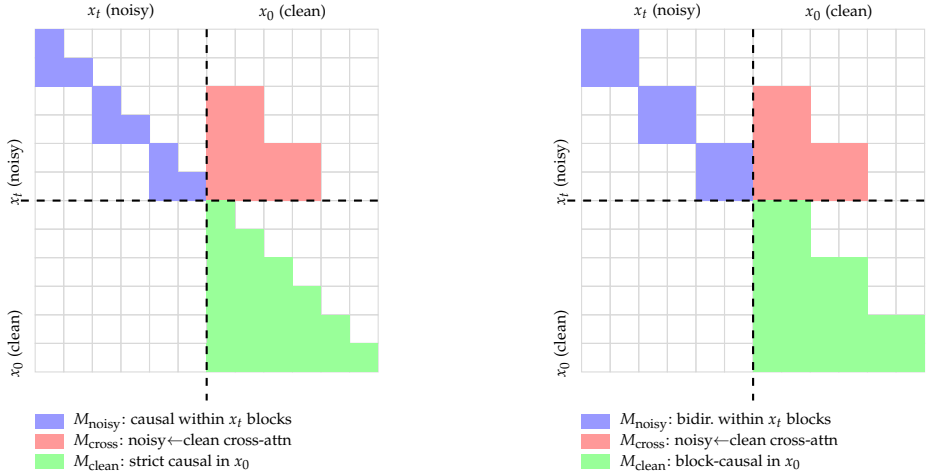
\begin{figure*}[t]
\centering
\begin{subfigure}[t]{0.48\textwidth}
\centering
\begin{tikzpicture}[scale=0.38, every node/.style={font=\tiny}]

    \foreach \i in {0,...,11} {
        \foreach \j in {0,...,11} {
            \draw[gray!30] (\j, -\i) rectangle (\j+1, -\i+1);
        }
    }

    \fill[blue!40] (0,0) rectangle (1,1);
    \fill[blue!40] (0,-1) rectangle (1,0);
    \fill[blue!40] (1,-1) rectangle (2,0);
    \fill[blue!40] (2,-2) rectangle (3,-1);
    \fill[blue!40] (2,-3) rectangle (3,-2);
    \fill[blue!40] (3,-3) rectangle (4,-2);
    \fill[blue!40] (4,-4) rectangle (5,-3);
    \fill[blue!40] (4,-5) rectangle (5,-4);
    \fill[blue!40] (5,-5) rectangle (6,-4);

    \fill[red!40] (6,-2) rectangle (8,-1);
    \fill[red!40] (6,-3) rectangle (8,-2);
    \fill[red!40] (6,-4) rectangle (10,-3);
    \fill[red!40] (6,-5) rectangle (10,-4);

    \fill[green!40] (6,-6) rectangle (7,-5);
    \fill[green!40] (6,-7) rectangle (8,-6);
    \fill[green!40] (6,-8) rectangle (9,-7);
    \fill[green!40] (6,-9) rectangle (10,-8);
    \fill[green!40] (6,-10) rectangle (11,-9);
    \fill[green!40] (6,-11) rectangle (12,-10);

    \draw[thick, dashed] (6, 1) -- (6, -12);
    \draw[thick, dashed] (-0.5, -5) -- (12, -5);

    \node[above] at (3, 1) {$x_t$ (noisy)};
    \node[above] at (9, 1) {$x_0$ (clean)};
    \node[left, rotate=90] at (-0.7, -2.5) {$x_t$ (noisy)};
    \node[left, rotate=90] at (-0.7, -8.5) {$x_0$ (clean)};

    \fill[blue!40] (0, -13) rectangle (1, -12.5);
    \node[right] at (1, -12.75) {$M_\text{noisy}$: causal within $x_t$ blocks};
    \fill[red!40] (0, -13.7) rectangle (1, -13.2);
    \node[right] at (1, -13.45) {$M_\text{cross}$: noisy$\leftarrow$clean cross-attn};
    \fill[green!40] (0, -14.4) rectangle (1, -13.9);
    \node[right] at (1, -14.15) {$M_\text{clean}$: strict causal in $x_0$};
\end{tikzpicture}
\caption{\textbf{I-DLM (Ours).} Strict causal within $x_t$ blocks and strict causal in $x_0$.}
\label{fig:attn_mask_ours}
\end{subfigure}
\hfill
\begin{subfigure}[t]{0.48\textwidth}
\centering
\begin{tikzpicture}[scale=0.38, every node/.style={font=\tiny}]

    \foreach \i in {0,...,11} {
        \foreach \j in {0,...,11} {
            \draw[gray!30] (\j, -\i) rectangle (\j+1, -\i+1);
        }
    }

    \fill[blue!40] (0,0) rectangle (2,1);
    \fill[blue!40] (0,-1) rectangle (2,0);
    \fill[blue!40] (2,-2) rectangle (4,-1);
    \fill[blue!40] (2,-3) rectangle (4,-2);
    \fill[blue!40] (4,-4) rectangle (6,-3);
    \fill[blue!40] (4,-5) rectangle (6,-4);

    \fill[red!40] (6,-2) rectangle (8,-1);
    \fill[red!40] (6,-3) rectangle (8,-2);
    \fill[red!40] (6,-4) rectangle (10,-3);
    \fill[red!40] (6,-5) rectangle (10,-4);

    \fill[green!40] (6,-6) rectangle (8,-5);
    \fill[green!40] (6,-7) rectangle (8,-6);
    \fill[green!40] (6,-8) rectangle (10,-7);
    \fill[green!40] (6,-9) rectangle (10,-8);
    \fill[green!40] (6,-10) rectangle (12,-9);
    \fill[green!40] (6,-11) rectangle (12,-10);

    \draw[thick, dashed] (6, 1) -- (6, -12);
    \draw[thick, dashed] (-0.5, -5) -- (12, -5);

    \node[above] at (3, 1) {$x_t$ (noisy)};
    \node[above] at (9, 1) {$x_0$ (clean)};
    \node[left, rotate=90] at (-0.7, -2.5) {$x_t$ (noisy)};
    \node[left, rotate=90] at (-0.7, -8.5) {$x_0$ (clean)};

    \fill[blue!40] (0, -13) rectangle (1, -12.5);
    \node[right] at (1, -12.75) {$M_\text{noisy}$: bidir.\ within $x_t$ blocks};
    \fill[red!40] (0, -13.7) rectangle (1, -13.2);
    \node[right] at (1, -13.45) {$M_\text{cross}$: noisy$\leftarrow$clean cross-attn};
    \fill[green!40] (0, -14.4) rectangle (1, -13.9);
    \node[right] at (1, -14.15) {$M_\text{clean}$: block-causal in $x_0$};
\end{tikzpicture}
\caption{\textbf{SDAR (Block Diffusion).} Bidirectional within $x_t$ blocks and block-causal in $x_0$.}
\label{fig:attn_mask_sdar}
\end{subfigure}
\caption{\textbf{Attention mask comparison} for block size $N{=}2$, sequence length $L{=}6$. Input is $[x_t \,|\, x_0]$ (noisy $\|$ clean). Rows are query positions; columns are key positions. Our I-DLM (left) uses strict causal attention everywhere, preserving AR compatibility. SDAR (right) uses bidirectional attention within noisy blocks and block-causal attention in the clean region. The three mask components---$M_\text{noisy}$ (noisy self-attention), $M_\text{cross}$ (noisy$\leftarrow$clean cross-attention), $M_\text{clean}$ (clean self-attention)---are color-coded.}
\label{fig:attn_mask}
\end{figure*}

Figure~\ref{fig:attn_mask} visualizes the attention mask used during I-DLM training, contrasted with SDAR's block-causal mask. The input sequence is the concatenation $[x_t \,|\, x_0]$, where $x_t$ is the all-masked (noisy) region and $x_0$ is the clean reference. The mask is composed of three components:

\begin{denseitemize}
    \item \textbf{Noisy self-attention ($M_\text{noisy}$):} Self-attention within the noisy region $x_t$. In our setting (\texttt{use\_regular\_causal=True}), this is \emph{causal within each block}: position $j$ in block $b$ attends only to positions $i \leq j$ in the same block. SDAR instead uses bidirectional attention within blocks.
    \item \textbf{Cross-attention ($M_\text{cross}$):} Cross-attention from noisy tokens to clean tokens. Each noisy block $b$ attends to clean tokens from all \emph{preceding} blocks ($b' < b$), providing conditional context from the clean reference.
    \item \textbf{Clean self-attention ($M_\text{clean}$):} Self-attention within the clean region $x_0$. In our setting, this is \emph{strict token-level causal} ($q \geq kv$), preserving the AR attention pattern exactly. SDAR instead uses block-causal attention here.
\end{denseitemize}

The final mask is $M = M_\text{noisy} \cup M_\text{cross} \cup M_\text{clean}$.

\section{ISD Step-by-Step Illustration}
\label{app:isd_steps}

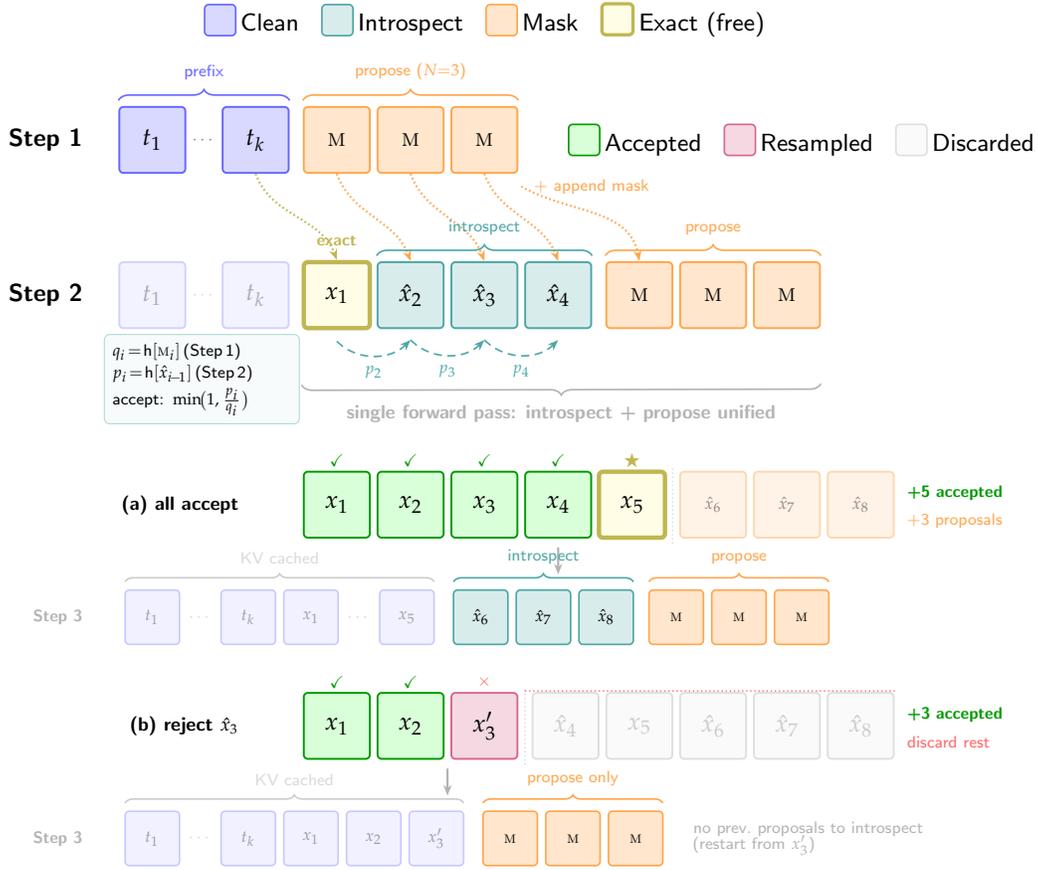
\begin{figure*}[t]
\centering
\resizebox{\textwidth}{!}{%
\begin{tikzpicture}[
    tokbase/.style={
        minimum width=0.85cm, minimum height=0.85cm,
        line width=0.7pt, font=\footnotesize\sffamily,
        inner sep=0pt, rounded corners=1.5pt,
    },
    smalltok/.style={
        minimum width=0.70cm, minimum height=0.70cm,
        line width=0.6pt, font=\tiny\sffamily,
        inner sep=0pt, rounded corners=1.5pt,
    },
    masktok/.style={tokbase, draw=orange!70, fill=orange!20},
    cleantok/.style={tokbase, draw=blue!60, fill=blue!15},
    verifytok/.style={tokbase, draw=teal!70, fill=teal!15},
    accepttok/.style={tokbase, draw=green!60!black, fill=green!15},
    exacttok/.style={tokbase, draw=yellow!70!black, fill=yellow!12, line width=1.6pt},
    resampletok/.style={tokbase, draw=purple!60, fill=purple!15},
    fadetok/.style={tokbase, draw=gray!25, fill=gray!4, text=gray!40},
    smasktok/.style={smalltok, draw=orange!70, fill=orange!20},
    scleantok/.style={smalltok, draw=blue!60, fill=blue!15, opacity=0.35},
    sverifytok/.style={smalltok, draw=teal!70, fill=teal!15},
    steplabel/.style={font=\small\sffamily\bfseries, text=black},
    annot/.style={font=\tiny\sffamily, text=gray!60},
    arr/.style={-{Stealth[length=4pt]}, line width=0.6pt, gray!60},
    brace/.style={decorate, decoration={brace, amplitude=4pt, raise=2pt}, line width=0.6pt},
    bracebelow/.style={decorate, decoration={brace, amplitude=4pt, raise=2pt, mirror}, line width=0.6pt},
]

\def\sp{0.95}

\newcommand{\lsw}[2]{\raisebox{-2pt}{\tikz\fill[#2, draw=#1, line width=0.6pt, rounded corners=1.5pt] (0,0) rectangle (0.4cm,0.4cm);}}
\newcommand{\lswt}[2]{\raisebox{-2pt}{\tikz\fill[#2, draw=#1, line width=1.2pt, rounded corners=1.5pt] (0,0) rectangle (0.4cm,0.4cm);}}

\node[font=\small\sffamily, anchor=south] at (5.0*\sp, 1.2) {%
    \lsw{blue!60}{blue!15}\,Clean\hskip8pt
    \lsw{teal!70}{teal!15}\,Introspect\hskip8pt
    \lsw{orange!70}{orange!20}\,Mask\hskip8pt
    \lswt{yellow!70!black}{yellow!12}\,Exact (free)%
};
\node[font=\small\sffamily, anchor=west] at (6.0*\sp, 0) {%
    \lsw{green!60!black}{green!15}\,Accepted\hskip8pt
    \lsw{purple!60}{purple!15}\,Resampled\hskip8pt
    \lsw{gray!25}{gray!4}\,Discarded%
};

\def\RA{0}
\node[steplabel, anchor=east] at (-0.3, \RA) {Step 1};

\node[cleantok] (r1t1) at (0.5*\sp, \RA) {$t_1$};
\node[font=\tiny\sffamily, text=gray] at (1.2*\sp, \RA) {$\cdots$};
\node[cleantok] (r1tk) at (1.9*\sp, \RA) {$t_k$};
\node[masktok] (r1m1) at (3.0*\sp, \RA) {\textsc{m}};
\node[masktok] (r1m2) at (4.0*\sp, \RA) {\textsc{m}};
\node[masktok] (r1m3) at (5.0*\sp, \RA) {\textsc{m}};

\draw[brace, blue!60] (r1t1.north west) -- (r1tk.north east)
    node[midway, above=6pt, annot, text=blue!60] {prefix};
\draw[brace, orange!70] (r1m1.north west) -- (r1m3.north east)
    node[midway, above=6pt, annot, text=orange!70] {propose ($N{=}3$)};

\def\RB{-2.0}
\node[steplabel, anchor=east] at (-0.3, \RB) {Step 2};

\node[cleantok, opacity=0.3] at (0.5*\sp, \RB) {$t_1$};
\node[font=\tiny\sffamily, text=gray, opacity=0.3] at (1.2*\sp, \RB) {$\cdots$};
\node[cleantok, opacity=0.3] at (1.9*\sp, \RB) {$t_k$};

\node[exacttok] (v1) at (3.0*\sp, \RB) {$x_1$};
\node[verifytok] (v2) at (4.0*\sp, \RB) {$\hat{x}_2$};
\node[verifytok] (v3) at (5.0*\sp, \RB) {$\hat{x}_3$};
\node[verifytok] (v4) at (6.0*\sp, \RB) {$\hat{x}_4$};

\node[masktok] (m1) at (7.1*\sp, \RB) {\textsc{m}};
\node[masktok] (m2) at (8.1*\sp, \RB) {\textsc{m}};
\node[masktok] (m3) at (9.1*\sp, \RB) {\textsc{m}};

\draw[arr, yellow!70!black, densely dotted, line width=0.8pt]
    (r1tk.south) to[out=-70, in=110] (v1.north);
\draw[arr, orange!70, densely dotted, line width=0.8pt]
    (r1m1.south) to[out=-70, in=110] (v2.north);
\draw[arr, orange!70, densely dotted, line width=0.8pt]
    (r1m2.south) to[out=-70, in=110] (v3.north);
\draw[arr, orange!70, densely dotted, line width=0.8pt]
    (r1m3.south) to[out=-70, in=110] (v4.north);

\draw[arr, orange!70, densely dotted, line width=0.8pt]
    (5.5*\sp, \RA - 0.6) to[out=-30, in=100]
    node[pos=0.0, right=1pt, annot, text=orange!70] {+ append mask}
    (m1.north);

\node[annot, text=yellow!70!black, font=\tiny\sffamily\bfseries, anchor=south] at (3.0*\sp, \RB+0.52) {exact};
\draw[brace, teal!70] (v2.north west) -- (v4.north east)
    node[midway, above=6pt, annot, text=teal!70] {introspect};
\draw[brace, orange!70] (m1.north west) -- (m3.north east)
    node[midway, above=6pt, annot, text=orange!70] {propose};

\draw[arr, teal!70, densely dashed]
    ([yshift=-4pt]v1.south) to[out=-45, in=-135]
    node[below=0pt, annot, text=teal!70] {$p_2$}
    ([yshift=-4pt]v2.south);
\draw[arr, teal!70, densely dashed]
    ([yshift=-4pt]v2.south) to[out=-45, in=-135]
    node[below=0pt, annot, text=teal!70] {$p_3$}
    ([yshift=-4pt]v3.south);
\draw[arr, teal!70, densely dashed]
    ([yshift=-4pt]v3.south) to[out=-45, in=-135]
    node[below=0pt, annot, text=teal!70] {$p_4$}
    ([yshift=-4pt]v4.south);

\node[draw=teal!30, fill=teal!3, rounded corners=2pt,
      inner sep=3pt, anchor=north east, font=\tiny\sffamily, text width=2.3cm, align=left]
    at (2.5*\sp, \RB-0.5) {%
    $q_i \!=\! \text{h}[\textsc{m}_i]$\,{\tiny(Step\,1)}\\[1pt]
    $p_i \!=\! \text{h}[\hat{x}_{i\!-\!1}]$\,{\tiny(Step\,2)}\\[1pt]
    accept: $\min\!\bigl(1,\tfrac{p_i}{q_i}\bigr)$
};

\draw[bracebelow, gray!60] ([yshift=-18pt]v1.south west) -- ([yshift=-18pt]m3.south east)
    node[midway, below=6pt, font=\scriptsize\sffamily\bfseries, text=gray!60] {single forward pass: introspect + propose unified};

\def\CA{-4.7}
\node[steplabel, anchor=east, font=\scriptsize\sffamily\bfseries] at (1.8*\sp, \CA) {(a) all accept};

\node[accepttok] at (3.0*\sp, \CA) {$x_1$};
\node[accepttok] at (4.0*\sp, \CA) {$x_2$};
\node[accepttok] at (5.0*\sp, \CA) {$x_3$};
\node[accepttok] at (6.0*\sp, \CA) {$x_4$};
\node[exacttok]  at (7.0*\sp, \CA) {$x_5$};

\draw[gray!30, densely dotted, line width=0.5pt]
    (7.55*\sp, \CA-0.5) -- (7.55*\sp, \CA+0.5);

\node[masktok, opacity=0.45, font=\tiny\sffamily] at (8.1*\sp, \CA) {$\hat{x}_6$};
\node[masktok, opacity=0.45, font=\tiny\sffamily] at (9.1*\sp, \CA) {$\hat{x}_7$};
\node[masktok, opacity=0.45, font=\tiny\sffamily] at (10.1*\sp, \CA) {$\hat{x}_8$};

\node[font=\tiny, text=green!60!black] at (3.0*\sp, \CA+0.57) {\checkmark};
\node[font=\tiny, text=green!60!black] at (4.0*\sp, \CA+0.57) {\checkmark};
\node[font=\tiny, text=green!60!black] at (5.0*\sp, \CA+0.57) {\checkmark};
\node[font=\tiny, text=green!60!black] at (6.0*\sp, \CA+0.57) {\checkmark};
\node[font=\tiny\sffamily\bfseries, text=yellow!70!black] at (7.0*\sp, \CA+0.57) {$\bigstar$};

\node[annot, font=\tiny\sffamily\bfseries, text=green!60!black, anchor=west] at (10.6*\sp, \CA+0.15) {+5 accepted};
\node[annot, text=orange!70, anchor=west] at (10.6*\sp, \CA-0.2) {+3 proposals};

\draw[arr, line width=0.8pt] (6.0*\sp, \CA-0.55) -- (6.0*\sp, \CA-0.9);

\def\RAA{\CA - 1.45}
\node[annot, anchor=east, font=\tiny\sffamily\bfseries] at (-0.3, \RAA) {Step 3};

\node[scleantok] at (0.5*\sp, \RAA) {\tiny$t_1$};
\node[font=\tiny\sffamily, text=gray, opacity=0.3] at (1.15*\sp, \RAA) {$\cdots$};
\node[scleantok] at (1.8*\sp, \RAA) {\tiny$t_k$};
\node[scleantok] at (2.65*\sp, \RAA) {\tiny$x_1$};
\node[font=\tiny\sffamily, text=gray, opacity=0.3] at (3.3*\sp, \RAA) {$\cdots$};
\node[scleantok] at (3.95*\sp, \RAA) {\tiny$x_5$};

\node[sverifytok] at (4.95*\sp, \RAA) {\tiny$\hat{x}_6$};
\node[sverifytok] at (5.8*\sp, \RAA) {\tiny$\hat{x}_7$};
\node[sverifytok] at (6.65*\sp, \RAA) {\tiny$\hat{x}_8$};

\node[smasktok] at (7.6*\sp, \RAA) {\tiny\textsc{m}};
\node[smasktok] at (8.45*\sp, \RAA) {\tiny\textsc{m}};
\node[smasktok] at (9.3*\sp, \RAA) {\tiny\textsc{m}};

\draw[brace, gray!40, decoration={amplitude=3pt, raise=1pt}]
    (0.5*\sp-0.35, \RAA+0.42) -- (3.95*\sp+0.35, \RAA+0.42)
    node[midway, above=4pt, annot, opacity=0.5] {\tiny KV cached};
\draw[brace, teal!70, decoration={amplitude=3pt, raise=1pt}]
    (4.95*\sp-0.35, \RAA+0.42) -- (6.65*\sp+0.35, \RAA+0.42)
    node[midway, above=4pt, annot, text=teal!70] {\tiny introspect};
\draw[brace, orange!70, decoration={amplitude=3pt, raise=1pt}]
    (7.6*\sp-0.35, \RAA+0.42) -- (9.3*\sp+0.35, \RAA+0.42)
    node[midway, above=4pt, annot, text=orange!70] {\tiny propose};

\def\CB{\RAA - 1.4}
\node[steplabel, anchor=east, font=\scriptsize\sffamily\bfseries] at (1.8*\sp, \CB) {(b) reject $\hat{x}_3$};

\node[accepttok] at (3.0*\sp, \CB) {$x_1$};
\node[accepttok] at (4.0*\sp, \CB) {$x_2$};
\node[resampletok] at (5.0*\sp, \CB) {$x'_3$};

\draw[gray!30, densely dotted, line width=0.5pt]
    (5.55*\sp, \CB-0.5) -- (5.55*\sp, \CB+0.5);

\node[fadetok] at (6.1*\sp, \CB) {$\hat{x}_4$};
\node[fadetok] at (7.1*\sp, \CB) {$x_5$};
\node[fadetok] at (8.1*\sp, \CB) {$\hat{x}_6$};
\node[fadetok] at (9.1*\sp, \CB) {$\hat{x}_7$};
\node[fadetok] at (10.1*\sp, \CB) {$\hat{x}_8$};

\node[font=\tiny, text=green!60!black] at (3.0*\sp, \CB+0.57) {\checkmark};
\node[font=\tiny, text=green!60!black] at (4.0*\sp, \CB+0.57) {\checkmark};
\node[font=\tiny, text=red!60] at (5.0*\sp, \CB+0.57) {$\times$};

\draw[red!60, line width=0.5pt, densely dotted]
    (5.55*\sp, \CB+0.45) -- (10.6*\sp, \CB+0.45);

\node[annot, font=\tiny\sffamily\bfseries, text=green!60!black, anchor=west] at (10.6*\sp, \CB+0.15) {+3 accepted};
\node[annot, text=red!60, anchor=west] at (10.6*\sp, \CB-0.2) {discard rest};

\draw[arr, line width=0.8pt] (4.5*\sp, \CB-0.55) -- (4.5*\sp, \CB-0.9);

\def\RBB{\CB - 1.45}
\node[annot, anchor=east, font=\tiny\sffamily\bfseries] at (-0.3, \RBB) {Step 3};

\node[scleantok] at (0.5*\sp, \RBB) {\tiny$t_1$};
\node[font=\tiny\sffamily, text=gray, opacity=0.3] at (1.15*\sp, \RBB) {$\cdots$};
\node[scleantok] at (1.8*\sp, \RBB) {\tiny$t_k$};
\node[scleantok] at (2.65*\sp, \RBB) {\tiny$x_1$};
\node[scleantok] at (3.5*\sp, \RBB) {\tiny$x_2$};
\node[scleantok] at (4.35*\sp, \RBB) {\tiny$x'_3$};

\node[smasktok] at (5.35*\sp, \RBB) {\tiny\textsc{m}};
\node[smasktok] at (6.2*\sp, \RBB) {\tiny\textsc{m}};
\node[smasktok] at (7.05*\sp, \RBB) {\tiny\textsc{m}};

\draw[brace, gray!40, decoration={amplitude=3pt, raise=1pt}]
    (0.5*\sp-0.35, \RBB+0.42) -- (4.35*\sp+0.35, \RBB+0.42)
    node[midway, above=4pt, annot, opacity=0.5] {\tiny KV cached};
\draw[brace, orange!70, decoration={amplitude=3pt, raise=1pt}]
    (5.35*\sp-0.35, \RBB+0.42) -- (7.05*\sp+0.35, \RBB+0.42)
    node[midway, above=4pt, annot, text=orange!70] {\tiny propose only};

\node[annot, anchor=west, text width=3.5cm] at (7.7*\sp, \RBB) {no prev.\ proposals to introspect\\[-1pt](restart from $x'_3$)};

\end{tikzpicture}
}%
\caption{\textbf{Detailed ISD illustration at stride $N{=}3$.} \textbf{Step 1}: Bootstrap---append 3 \texttt{[MASK]} tokens, producing $x_1$ (exact) and proposals $\hat{x}_2, \hat{x}_3, \hat{x}_4$. \textbf{Step 2}: Single forward pass that introspects on previous proposals (computing causal anchors $p_k$) while generating new proposals. \textbf{(a) All accept}: 4 tokens accepted + bonus $x_5$; Step~3 introspects on new proposals. \textbf{(b) Reject $\hat{x}_3$}: $x_1, x_2$ accepted, $x'_3$ resampled, rest discarded; Step~3 is a pure propose step.}
\label{fig:isd_detailed}
\end{figure*}

Figure~\ref{fig:isd_detailed} provides a detailed step-by-step illustration of Introspective Strided Decoding at stride $N{=}3$, showing both the all-accept and rejection cases.

\section{Lossless ISD with Gated LoRA}
\label{app:gated_lora}

\begin{figure*}[t]
\centering
\begin{tikzpicture}[
    scale=1.0,
    tok/.style={minimum width=1.1cm, minimum height=0.7cm, draw, rounded corners=2pt, font=\scriptsize\sffamily, inner sep=1pt, line width=0.6pt},
    clean/.style={tok, draw=blue!60, fill=blue!10},
    verify/.style={tok, draw=teal!70, fill=teal!10},
    mask/.style={tok, draw=orange!70, fill=orange!15},
    exact/.style={tok, draw=yellow!70!black, fill=yellow!10, line width=1.2pt},
    layer/.style={minimum width=1.1cm, minimum height=0.55cm, rounded corners=2pt, font=\tiny\sffamily, inner sep=1pt},
    base/.style={layer, draw=gray!60, fill=gray!8},
    lora/.style={layer, draw=purple!60, fill=purple!10},
    arr/.style={-{Stealth[length=3pt]}, line width=0.5pt, gray!70},
    annot/.style={font=\tiny\sffamily, text=gray!70},
]

\def\sp{1.5}
\node[clean] (t1) at (0, 0) {$t_1$};
\node[font=\scriptsize, text=gray] at (1*\sp, 0) {$\cdots$};
\node[clean] (tk) at (2*\sp, 0) {$t_k$};
\node[verify] (v2) at (3*\sp, 0) {$\hat{x}_2$};
\node[verify] (v3) at (4*\sp, 0) {$\hat{x}_3$};
\node[exact] (x1) at (5*\sp, 0) {$x_1$};
\node[mask] (m1) at (6*\sp, 0) {\textsc{m}};
\node[mask] (m2) at (7*\sp, 0) {\textsc{m}};

\node[annot, above=2pt] at ($(t1.north)!0.5!(tk.north)$) {prefix};
\node[annot, above=2pt, text=teal!70] at ($(v2.north)!0.5!(v3.north)$) {introspect};
\node[annot, above=2pt] at (x1.north) {exact};
\node[annot, above=2pt, text=orange!70] at ($(m1.north)!0.5!(m2.north)$) {propose};

\foreach \n/\x in {t1/0, tk/2, v2/3, v3/4, x1/5} {
    \node[base] (b\x) at (\x*\sp, -1.1) {$Wx$};
    \draw[arr] (\n.south) -- (b\x.north);
}

\foreach \n/\x in {m1/6, m2/7} {
    \node[base, minimum width=0.5cm] (b\x) at (\x*\sp-0.45, -1.1) {\tiny$Wx$};
    \node[font=\tiny, text=purple!60] at (\x*\sp, -1.1) {$+$};
    \node[lora, minimum width=0.5cm] (l\x) at (\x*\sp+0.45, -1.1) {\tiny$ABx$};
    \draw[arr] (\n.south) -- (b\x.north);
    \draw[arr] (\n.south) -- (l\x.north);
}

\foreach \x/\lab/\col in {0/$p$/gray, 2/$p$/gray, 3/$p$/teal, 4/$p$/teal, 5/$p$/gray} {
    \node[font=\tiny\sffamily, text=\col] at (\x*\sp, -1.75) {\lab};
}
\foreach \x in {6,7} {
    \node[font=\tiny\sffamily, text=purple!70] at (\x*\sp, -1.75) {$q$};
}


\draw[decorate, decoration={brace, amplitude=3pt, mirror, raise=3pt}, line width=0.5pt, gray]
    (0-0.55, -1.95) -- (5*\sp+0.55, -1.95)
    node[midway, below=7pt, font=\tiny\sffamily, text=gray] {base-only weights $\Rightarrow$ $p$ = exact AR distribution};

\draw[decorate, decoration={brace, amplitude=3pt, mirror, raise=3pt}, line width=0.5pt, purple!60]
    (6*\sp-0.55, -1.95) -- (7*\sp+0.55, -1.95)
    node[midway, below=7pt, font=\tiny\sffamily, text=purple!60] {base+LoRA weights $\Rightarrow$ $q$ aligned with $p$};

\node[clean, minimum width=0.6cm, minimum height=0.4cm] at (0, -3.0) {};
\node[font=\tiny\sffamily, right] at (0.4, -3.0) {Clean};
\node[verify, minimum width=0.6cm, minimum height=0.4cm] at (1.5*\sp, -3.0) {};
\node[font=\tiny\sffamily, right] at (1.5*\sp+0.4, -3.0) {Introspect};
\node[exact, minimum width=0.6cm, minimum height=0.4cm] at (3*\sp, -3.0) {};
\node[font=\tiny\sffamily, right] at (3*\sp+0.4, -3.0) {Exact};
\node[mask, minimum width=0.6cm, minimum height=0.4cm] at (4.5*\sp, -3.0) {};
\node[font=\tiny\sffamily, right] at (4.5*\sp+0.4, -3.0) {Propose};
\node[base, minimum width=0.6cm, minimum height=0.4cm] at (6*\sp, -3.0) {};
\node[font=\tiny\sffamily, right] at (6*\sp+0.4, -3.0) {Base ($W$)};
\node[lora, minimum width=0.6cm, minimum height=0.4cm] at (7.2*\sp, -3.0) {};
\node[font=\tiny\sffamily, right] at (7.2*\sp+0.4, -3.0) {LoRA ($AB$)};

\end{tikzpicture}
\caption{\textbf{Gated LoRA in Residual ISD (R-ISD).} During a single forward pass, \texttt{[MASK]} (propose) positions compute $Wx + ABx$ using base+LoRA weights, producing proposal distributions $q$. Clean and introspect positions compute $Wx$ using base-only weights, producing the causal anchor distribution $p$---identical to a pure base AR forward pass. Because of causal attention, introspection positions never attend to \texttt{[MASK]} positions, so $p$ is computed entirely from base-only KV entries. This makes the output bit-for-bit lossless with respect to the base AR model.}
\label{fig:gated_lora}
\end{figure*}
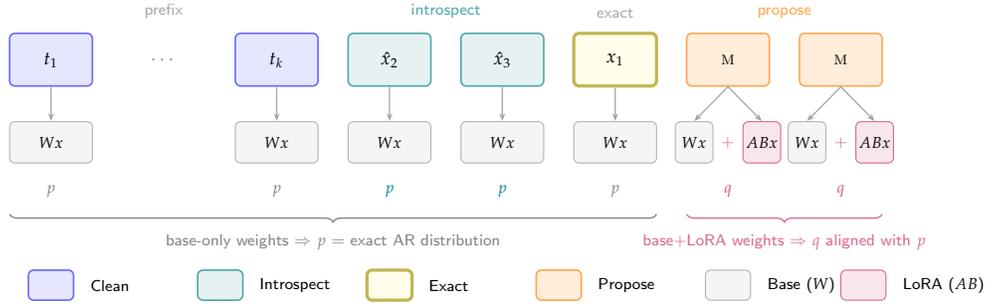

Figure~\ref{fig:gated_lora} illustrates the gated LoRA mechanism used in Residual ISD (R-ISD). During each forward pass, token positions are partitioned into two types based on their input: \texttt{[MASK]} positions (proposals) activate the LoRA residual, while clean positions (introspection) use base-model-only weights. Because of strict causal attention, introspection positions cannot attend to any \texttt{[MASK]} position---their KV cache entries are computed entirely from base weights over clean tokens.

\paragraph{Per-position computation.} Table~\ref{tab:lora_positions} summarizes the computation at each position type. The gated LoRA mechanism applies a per-token binary mask $\mathbf{m}$: each linear layer computes $h_j \gets Wx_j + \mathbf{1}_{[j \in \texttt{M}]} \cdot BA\,x_j$, where $\mathbf{1}_{[j \in \texttt{M}]}$ is $1$ for \texttt{[MASK]} positions and $0$ otherwise.

\begin{table}[h]
\centering
\caption{\textbf{Per-position breakdown in R-ISD.} Introspect and exact positions use base-only weights, producing the AR-identical causal anchor $p$. Propose positions add the LoRA residual to produce aligned proposals $q$.}
\label{tab:lora_positions}
\small
\begin{tabular}{llll}
\toprule
\textbf{Position} & \textbf{Weights} & \textbf{Output} & \textbf{Role} \\
\midrule
$\hat{x}_k$ (introspect) & $Wx$ (base only) & $p_k$ (causal anchor) & Produces $p$ for $p/q$ acceptance \\
$x_1$ (exact) & $Wx$ (base only) & $p$ (AR-identical) & Quality-guaranteed token \\
\texttt{[M]}$_k$ (propose) & $Wx + BAx$ (base+LoRA) & $q_k$ (proposal) & Strided proposal aligned with $p$ \\
\bottomrule
\end{tabular}
\end{table}

The overall R-ISD pipeline can be expressed compactly as:
\begin{equation}
\underbrace{x_{i+1}}_{\substack{\text{from } h[c_i]\\\text{(base only)}}}
,\;\;
\underbrace{\hat{x}_{i+2}, \ldots, \hat{x}_{i+N}}_{\substack{\text{from } h[\texttt{M}_{1:N-1}]\\
\text{(base+LoRA)}}}
= \pi_\theta(c_{1:i},\, \texttt{M}_{1:N-1})
\label{eq:risd_pipeline}
\end{equation}
where $\pi_\theta$ denotes the model with gated LoRA: $W \to W + \mathbf{m} \odot BA$. With LoRA active on \texttt{[MASK]} positions, all proposals become higher quality, increasing the introspective acceptance rate while the causal anchor $p$ remains the exact base AR distribution.

\section{Training, Serving, and Evaluation Details}
\label{app:training_details}
\label{app:eval_details}

\subsection{Training Setup}
\label{app:training_setup}

\textbf{Base models and data.}
We train two I-DLM variants: \textbf{I-DLM-8B} and \textbf{I-DLM-32B}, converted from Qwen3-8B and Qwen3-32B~\citep{yang2025qwen3}, respectively, using the introspective-consistency training recipe described in Section~\ref{sec:training}. Our training codebase builds on the open-source SDAR framework(https://github.com/JetAstra/SDAR). 
The training data consists of \textbf{4.5B tokens} of responses generated across reasoning datasets. All training is performed on \textbf{8 H100 GPUs}.


\textbf{Training schedule.}
For I-DLM-8B, we train for \textbf{2 epochs} with a stride curriculum: the first epoch trains with stride $N{=}2$, and the second epoch trains with stride $N{=}3$. During the first two stride expansions ($N{=}1 \to 2$ and $N{=}2 \to 3$), we use a fixed low scale of $0.2$ for the cross-entropy loss on clean tokens to speed up masked token learning. Starting from the $N{=}3 \to 4$ expansion, we switch to the auto-balanced loss scaling described in Eq.~\ref{eq:loss}.
For I-DLM-32B, we train for \textbf{1 epoch} with stride $N{=}2$ with LoRA using rank 1024 and a fixed scale of $0.2$. We use $N{=}4$ for evaluation without additional full training as well.

\textbf{Hyperparameters.}
We use full fine-tuning with DeepSpeed ZeRO Stage 2. The learning rate is $1 \times 10^{-5}$ with cosine decay and a warmup ratio of 0.03. We use a per-device batch size of 1 with gradient accumulation steps of 4, yielding an effective batch size of 32 across 8 GPUs. The maximum sequence length is 4096 tokens. Training uses bf16 mixed precision. 

\textbf{LoRA training.}
For lossless ISD (R-ISD), we additionally train LoRA adapters with \textbf{rank 128} on the same data using the same hyperameters with a learning rate of 2e-4. The inference follows the gated residual design described in Section~\ref{sec:inference}.

\subsection{Hardware}
All experiments are conducted on NVIDIA H100 80GB SXM GPUs with NVLink interconnect. We use CUDA 12.9 with FlashInfer for attention computation. CUDA graphs are enabled for all configurations.

\subsection{Serving Configuration}

Table~\ref{tab:serving_config} summarizes the serving configurations used across all models.

\begin{table}[h]
\centering
\caption{\textbf{Serving configurations.} All models are served with SGLang on H100 GPUs.}
\label{tab:serving_config}
\small
\setlength{\tabcolsep}{3pt}
\begin{tabular}{lccccc}
\toprule
\textbf{Model} & \textbf{TP} & \textbf{\#Servers} & \textbf{Dtype} & \textbf{DLLM Config} & \textbf{LoRA} \\
\midrule
\multicolumn{6}{l}{\textit{8B models (1 GPU per server)}} \\
Qwen3-8B (AR) & 1 & 8 & bf16 & --- & --- \\
I-DLM-8B ($N{=}3$) & 1 & 8 & bf16 & blockN3 & --- \\
I-DLM-8B ($N{=}5$) & 1 & 8 & bf16 & blockN5 & --- \\
I-DLM-8B (Lossless) & 1 & 8 & bf16 & blockN3+LoRA & r=128 \\
EAGLE-3 & 1 & 8 & bf16 & --- & --- \\
\midrule
\multicolumn{6}{l}{\textit{32B models (2 GPUs per server)}} \\
Qwen3-32B (AR) & 2 & 4 & bf16 & --- & --- \\
I-DLM-32B ($N{=}3$) & 2 & 4 & bf16 & blockN3 & --- \\
I-DLM-32B (Lossless) & 2 & 4 & bf16 & blockN3+LoRA & r=1024 \\
\bottomrule
\end{tabular}
\end{table}

\paragraph{ISD algorithm configuration.}
Table~\ref{tab:isd_config} lists the ISD algorithm parameters used for each stride configuration.

\begin{table}[h]
\centering
\caption{\textbf{ISD algorithm configurations.} ``blockN$k$'' denotes stride $N{=}k$ with $\text{block\_size}=2k{-}1$.}
\label{tab:isd_config}
\small
\begin{tabular}{lcccccc}
\toprule
\textbf{Config} & \textbf{$N$} & \textbf{block\_size} & \textbf{temperature} & \textbf{top\_$k$} & \textbf{top\_$p$} & \textbf{spec\_verify} \\
\midrule
blockN3 & 3 & 5 & 1.0 & 50 & 0.95 & true \\
blockN5 & 5 & 9 & 1.0 & 50 & 0.95 & true \\
blockN3+LoRA & 3 & 5 & 1.0 & 50 & 0.95 & true \\
\bottomrule
\end{tabular}
\end{table}

\paragraph{Speculative decoding baselines.}
Table~\ref{tab:spec_baselines} details the configurations for speculative decoding baselines. All baselines use Qwen3-8B as the target model and are served with SGLang.

\begin{table}[h]
\centering
\caption{\textbf{Speculative decoding baseline configurations.} ``Steps'' and ``topk'' are EAGLE-3 draft parameters.}
\label{tab:spec_baselines}
\small
\setlength{\tabcolsep}{3pt}
\begin{tabular}{llcl}
\toprule
\textbf{Method} & \textbf{Draft Model} & \textbf{Draft Params} & \textbf{Notes} \\
\midrule
EAGLE-3 & Tengyunw/qwen3\_8b\_eagle3 & steps=3, topk=1, d=4 & 3 specs verified per step \\
\bottomrule
\end{tabular}
\end{table}

\paragraph{SGLang server parameters.}
Common parameters across all configurations: \texttt{mem-fraction-static=0.85}, \texttt{attention-backend=flashinfer}, \texttt{disable-radix-cache=true} (required for DLLM KV trim). CUDA graph capture is enabled by default; the DLLM extend forward (all $2N{-}1$ tokens) is captured into a single graph per batch size, with attention metadata updated via in-place tensor writes before each replay.

\subsection{Evaluation Configuration}

Table~\ref{tab:eval_config} details the evaluation settings for each benchmark.

\begin{table}[h]
\centering
\caption{\textbf{Evaluation configurations per benchmark.} All benchmarks use the Qwen3 chat template with thinking mode enabled. ``max\_tokens'' controls the maximum generation length including the \texttt{<think>} block.}
\label{tab:eval_config}
\small
\setlength{\tabcolsep}{3pt}
\begin{tabular}{llccl}
\toprule
\textbf{Category} & \textbf{Benchmark} & \textbf{\#Problems} & \textbf{max\_tokens} & \textbf{Metric / Extraction} \\
\midrule
\multirow{5}{*}{Knowledge}
 & ARC-C & 1,172 & 32,768 & Accuracy; choice letter \\
 & MMLU & 14,042 & 32,768 & Accuracy; ABCD extraction \\
 & MMLU-Pro & 12,032 & 32,768 & Accuracy; ABCD extraction \\
 & GPQA-Diamond & 198 & 32,768 & Accuracy; ABCD extraction \\
 & GPQA & 448 & 32,768 & Accuracy; ABCD extraction \\
\midrule
\multirow{5}{*}{Math}
 & GSM8K & 1,319 & 32,768 & Accuracy; \texttt{\textbackslash boxed\{\}} \\
 & MATH-500 & 500 & 32,768 & Accuracy; \texttt{\textbackslash boxed\{\}} \\
 & MathBench & 3,709 & 32,768 & Accuracy; numerical \\
 & AIME-24 & 30 & 32,768 & Accuracy; \texttt{\textbackslash boxed\{\}} \\
 & AIME-25 & 30 & 32,768 & Accuracy; \texttt{\textbackslash boxed\{\}} \\
\midrule
\multirow{3}{*}{Code}
 & HumanEval & 164 & 32,768 & pass@1; code execution \\
 & MBPP (sanitized) & 257 & 32,768 & pass@1; code execution \\
 & LCB-v6 & 175 & 32,768 & pass@1; code execution \\
\midrule
Instruction & IFEval & 541 & 32,768 & Prompt-strict accuracy \\
\midrule
Knowledge & TriviaQA & 17,944 & 32,768 & Exact match \\
\bottomrule
\end{tabular}
\end{table}

\paragraph{Sampling parameters.}
For quality evaluation, all models use temperature $t{=}1.0$, top-$k{=}50$, top-$p{=}0.95$ (matching the ISD algorithm configuration). For the AR baseline (Qwen3-8B/32B), the same sampling parameters are used to ensure a fair comparison. For each benchmark, results are averaged over three runs.

\paragraph{Answer extraction.}
For math benchmarks (GSM8K, MATH-500, AIME), we extract the final \texttt{\textbackslash boxed\{...\}} answer after stripping the \texttt{<think>...</think>} block. For multiple-choice benchmarks (ARC-C, MMLU, MMLU-Pro, GPQA), we extract the answer letter using the pattern \texttt{ANSWER: [A-D]}. For code benchmarks (HumanEval, MBPP, LCB), we extract the last Python code block containing a function definition and execute it against the provided test cases using a sandboxed subprocess with a 10-second timeout. For IFEval, we use the official Google evaluator after stripping thinking blocks.

\paragraph{Baseline reproduction.}
Results for LLaDA-2.1-mini are reproduced using SGLang with the official configuration (\texttt{block\_size=4}, \texttt{threshold=0.95}, \texttt{edit\_threshold=0.9}). Results for SDAR are reproduced using SGLang's DLLM serving mode. Results for EAGLE-3 are reproduced using its SGLang integration. All other baseline results are taken from their original publications.

\subsection{Throughput Benchmark Configuration}

For latency and throughput measurements (Section~\ref{fig:throughput}), we use genai-bench\footnote{\url{https://github.com/sgl-project/genai-bench}} with the following settings:

\begin{table}[h]
\centering
\caption{\textbf{Throughput benchmark configuration.}}
\label{tab:throughput_config}
\small
\begin{tabular}{ll}
\toprule
\textbf{Parameter} & \textbf{Value} \\
\midrule
Concurrency levels & $C \in \{1, 2, 4, 8, 16, 32, 48, 64\}$ \\
Request arrival & Burst mode (all requests submitted simultaneously) \\
Output length & 2,048 tokens (fixed) \\
Measurement & Total tokens / wall-clock time (first to last completion) \\
Warmup & 5 requests discarded before measurement \\
GPU & 1$\times$ H100 (TP=1) for 8B; 2$\times$ H100 (TP=2) for 32B \\
\bottomrule
\end{tabular}
\end{table}

\subsection{Infrastructure Ablation Configuration}

The ablation study in Section~\ref{fig:ablation} isolates the contribution of each serving optimization by disabling them individually via environment variables. Table~\ref{tab:ablation_config} lists the ablation toggles.

\begin{table}[h]
\centering
\caption{\textbf{Infrastructure ablation toggles.} Each toggle disables one optimization to measure its isolated contribution.}
\label{tab:ablation_config}
\small
\setlength{\tabcolsep}{3pt}
\begin{tabular}{lll}
\toprule
\textbf{Toggle} & \textbf{Disables} & \textbf{Effect} \\
\midrule
\texttt{-{}-disable-cuda-graph} & CUDA graph capture/replay & Forces eager model forward \\
\texttt{NO\_DECODE\_LOOP=1} & Stationary-batch decode loop & Falls back to full scheduler pipeline \\
\texttt{NO\_ARGMAX\_PROPOSALS=1} & Argmax for draft positions & Uses full sampling for specs \\
\texttt{NO\_FUSED\_VERIFY=1} & Fused Triton verify kernel & Uses 7+ separate kernel launches \\
\texttt{NO\_DEFERRED\_STREAM=1} & Deferred stream\_output & Runs on critical path \\
\texttt{DLLM\_USE\_RAGGED=1} & Paged-only attention & Forces cascade (3 kernels/layer) \\
\bottomrule
\end{tabular}
\end{table}

\noindent The ablation is cumulative: starting from the naive configuration (all optimizations disabled), we add each optimization one at a time and measure throughput at $C{=}1$, $C{=}8$, and $C{=}32$ on a single H100.

\section{Additional Results}
\label{app:results}

\subsection{Peak Throughput on Different Hardware}

Table~\ref{tab:peak_tps} reports peak per-request TPS under favorable conditions (low concurrency, long generation), using different model variants and stride configurations. The base I-DLM-8B is trained at $N{=}3$; the $N{=}4$ checkpoint is obtained via stride extension training from I-DLM-8B, and $N{=}8$ is further extended from $N{=}4$. The LoRA variant uses a rank-128 adapter with segment-gated conditional activation (R-ISD).

\begin{table}[h]
\centering
\caption{\textbf{Peak per-request TPS} at various stride, hardware, and model configurations. All models are 8B scale.}
\label{tab:peak_tps}
\small
\begin{tabular}{llcccc}
\toprule
\textbf{Hardware} & \textbf{TP} & \textbf{$N$} & \textbf{Model Variant} & \textbf{Benchmark} & \textbf{TPS} \\
\midrule
B200 & 2 & 8 & I-DLM-8B (ext.\ training $N{=}8$) & GSM8K & 925 \\
H100 & 2 & 8 & I-DLM-8B (ext.\ training $N{=}8$) & MBPP & 685 \\
H100 & 1 & 4 & I-DLM-8B (ext.\ training $N{=}4$) & MATH-500 & 341 \\
H100 & 1 & 4 & I-DLM-8B (ext.\ training $N{=}4$) & AIME & 314 \\
H100 & 1 & 4 & I-DLM-8B (ext.\ training $N{=}4$) & ShareGPT & 319 \\
H100 & 1 & 3 & I-DLM-8B & MATH-500 & 272 \\
H100 & 1 & 3 & I-DLM-8B R-ISD (LoRA $r{=}128$) & MATH-500 & 240 \\
\bottomrule
\end{tabular}
\end{table}

\label{sec:other_system_opt}

%% file: colm2026_conference.bib
@article{yang2025qwen3,
  title={Qwen3 technical report},
  author={Yang, An and Li, Anfeng and Yang, Baosong and Zhang, Beichen and Hui, Binyuan and Zheng, Bo and Yu, Bowen and Gao, Chang and Huang, Chengen and Lv, Chenxu and others},
  journal={arXiv preprint arXiv:2505.09388},
  year={2025}
}

@article{austin2021structured,
  title={Structured denoising diffusion models in discrete state-spaces},
  author={Austin, Jacob and Johnson, Daniel D and Ho, Jonathan and Tarlow, Daniel and Van Den Berg, Rianne},
  journal={Advances in neural information processing systems},
  volume={34},
  pages={17981--17993},
  year={2021}
}

@article{sahoo2024simple,
  title={Simple and effective masked diffusion language models},
  author={Sahoo, Subham S and Arriola, Marianne and Schiff, Yair and Gokaslan, Aaron and Marroquin, Edgar and Chiu, Justin T and Rush, Alexander and Kuleshov, Volodymyr},
  journal={Advances in Neural Information Processing Systems},
  volume={37},
  pages={130136--130184},
  year={2024}
}

@article{nie2025large,
  title={Large language diffusion models},
  author={Nie, Shen and Zhu, Fengqi and You, Zebin and Zhang, Xiaolu and Ou, Jingyang and Hu, Jun and Zhou, Jun and Lin, Yankai and Wen, Ji-Rong and Li, Chongxuan},
  journal={arXiv preprint arXiv:2502.09992},
  year={2025}
}

@article{wu2025fast,
  title={Fast-dllm: Training-free acceleration of diffusion llm by enabling kv cache and parallel decoding},
  author={Wu, Chengyue and Zhang, Hao and Xue, Shuchen and Liu, Zhijian and Diao, Shizhe and Zhu, Ligeng and Luo, Ping and Han, Song and Xie, Enze},
  journal={arXiv preprint arXiv:2505.22618},
  year={2025}
}

@article{liu2025wedlm,
  title={Wedlm: Reconciling diffusion language models with standard causal attention for fast inference},
  author={Liu, Aiwei and He, Minghua and Zeng, Shaoxun and Zhang, Sijun and Zhang, Linhao and Wu, Chuhan and Jia, Wei and Liu, Yuan and Zhou, Xiao and Zhou, Jie},
  journal={arXiv preprint arXiv:2512.22737},
  year={2025}
}

@article{gloeckle2024better,
  title={Better \& faster large language models via multi-token prediction},
  author={Gloeckle, Fabian and Idrissi, Badr Youbi and Rozi{\`e}re, Baptiste and Lopez-Paz, David and Synnaeve, Gabriel},
  journal={arXiv preprint arXiv:2404.19737},
  year={2024}
}

@inproceedings{leviathan2023fast,
  title={Fast inference from transformers via speculative decoding},
  author={Leviathan, Yaniv and Kalman, Matan and Matias, Yossi},
  booktitle={International Conference on Machine Learning},
  pages={19274--19286},
  year={2023},
  organization={PMLR}
}

@article{li2025eagle,
  title={Eagle-3: Scaling up inference acceleration of large language models via training-time test},
  author={Li, Yuhui and Wei, Fangyun and Zhang, Chao and Zhang, Hongyang},
  journal={arXiv preprint arXiv:2503.01840},
  year={2025}
}

@inproceedings{kou2024cllms,
  title={Cllms: Consistency large language models},
  author={Kou, Siqi and Hu, Lanxiang and He, Zhezhi and Deng, Zhijie and Zhang, Hao},
  booktitle={Forty-first International Conference on Machine Learning},
  year={2024}
}

@article{cheng2025sdar,
  title={Sdar: A synergistic diffusion-autoregression paradigm for scalable sequence generation},
  author={Cheng, Shuang and Bian, Yihan and Liu, Dawei and Zhang, Linfeng and Yao, Qian and Tian, Zhongbo and Wang, Wenhai and Guo, Qipeng and Chen, Kai and Qi, Biqing and others},
  journal={arXiv preprint arXiv:2510.06303},
  year={2025}
}

@article{tian2025next,
  title={From next-token to next-block: A principled adaptation path for diffusion llms},
  author={Tian, Yuchuan and Liang, Yuchen and Zhang, Shuo and Shu, Yingte and Yang, Guangwen and He, Wei and Fang, Sibo and Guo, Tianyu and Han, Kai and Xu, Chao and others},
  journal={arXiv preprint arXiv:2512.06776},
  year={2025}
}

@article{lou2023discrete,
  title={Discrete diffusion language modeling by estimating the ratios of the data distribution},
  author={Lou, Aaron and Meng, Chenlin and Ermon, Stefano},
  year={2023}
}

@article{bie2025llada2,
  title={Llada2. 0: Scaling up diffusion language models to 100b},
  author={Bie, Tiwei and Cao, Maosong and Chen, Kun and Du, Lun and Gong, Mingliang and Gong, Zhuochen and Gu, Yanmei and Hu, Jiaqi and Huang, Zenan and Lan, Zhenzhong and others},
  journal={arXiv preprint arXiv:2512.15745},
  year={2025}
}

@article{bie2026llada2,
  title={Llada2. 1: Speeding up text diffusion via token editing},
  author={Bie, Tiwei and Cao, Maosong and Cao, Xiang and Chen, Bingsen and Chen, Fuyuan and Chen, Kun and Du, Lun and Feng, Daozhuo and Feng, Haibo and Gong, Mingliang and others},
  journal={arXiv preprint arXiv:2602.08676},
  year={2026}
}

@article{ye2025dream,
  title={Dream 7b: Diffusion large language models},
  author={Ye, Jiacheng and Xie, Zhihui and Zheng, Lin and Gao, Jiahui and Wu, Zirui and Jiang, Xin and Li, Zhenguo and Kong, Lingpeng},
  journal={arXiv preprint arXiv:2508.15487},
  year={2025}
}

@article{arriola2025block,
  title={Block diffusion: Interpolating between autoregressive and diffusion language models},
  author={Arriola, Marianne and Gokaslan, Aaron and Chiu, Justin T and Yang, Zhihan and Qi, Zhixuan and Han, Jiaqi and Sahoo, Subham Sekhar and Kuleshov, Volodymyr},
  journal={arXiv preprint arXiv:2503.09573},
  year={2025}
}

@article{gong2024scaling,
  title={Scaling diffusion language models via adaptation from autoregressive models},
  author={Gong, Shansan and Agarwal, Shivam and Zhang, Yizhe and Ye, Jiacheng and Zheng, Lin and Li, Mukai and An, Chenxin and Zhao, Peilin and Bi, Wei and Han, Jiawei and others},
  journal={arXiv preprint arXiv:2410.17891},
  year={2024}
}

@article{deschenaux2024beyond,
  title={Beyond autoregression: Fast llms via self-distillation through time},
  author={Deschenaux, Justin and Gulcehre, Caglar},
  journal={arXiv preprint arXiv:2410.21035},
  year={2024}
}

@article{fu2025efficient,
  title={Efficient-dlm: From autoregressive to diffusion language models, and beyond in speed},
  author={Fu, Yonggan and Whalen, Lexington and Ye, Zhifan and Dong, Xin and Diao, Shizhe and Liu, Jingyu and Wu, Chengyue and Zhang, Hao and Xie, Enze and Han, Song and others},
  journal={arXiv preprint arXiv:2512.14067},
  year={2025}
}

@article{liu2025tidar,
  title={Tidar: Think in diffusion, talk in autoregression},
  author={Liu, Jingyu and Dong, Xin and Ye, Zhifan and Mehta, Rishabh and Fu, Yonggan and Singh, Vartika and Kautz, Jan and Zhang, Ce and Molchanov, Pavlo},
  journal={arXiv preprint arXiv:2511.08923},
  year={2025}
}

@article{chen2023accelerating,
  title={Accelerating large language model decoding with speculative sampling},
  author={Chen, Charlie and Borgeaud, Sebastian and Irving, Geoffrey and Lespiau, Jean-Baptiste and Sifre, Laurent and Jumper, John},
  journal={arXiv preprint arXiv:2302.01318},
  year={2023}
}

@article{cai2024medusa,
  title={Medusa: Simple llm inference acceleration framework with multiple decoding heads},
  author={Cai, Tianle and Li, Yuhong and Geng, Zhengyang and Peng, Hongwu and Lee, Jason D and Chen, Deming and Dao, Tri},
  journal={arXiv preprint arXiv:2401.10774},
  year={2024}
}

@article{li2024eagle,
  title={Eagle: Speculative sampling requires rethinking feature uncertainty},
  author={Li, Yuhui and Wei, Fangyun and Zhang, Chao and Zhang, Hongyang},
  journal={arXiv preprint arXiv:2401.15077},
  year={2024}
}

@article{miao2023specinfer,
  title={Specinfer: Accelerating generative large language model serving with tree-based speculative inference and verification},
  author={Miao, Xupeng and Oliaro, Gabriele and Zhang, Zhihao and Cheng, Xinhao and Wang, Zeyu and Zhang, Zhengxin and Wong, Rae Ying Yee and Zhu, Alan and Yang, Lijie and Shi, Xiaoxiang and others},
  journal={arXiv preprint arXiv:2305.09781},
  year={2023}
}

@article{samragh2025your,
  title={Your llm knows the future: Uncovering its multi-token prediction potential},
  author={Samragh, Mohammad and Kundu, Arnav and Harrison, David and Nishu, Kumari and Naik, Devang and Cho, Minsik and Farajtabar, Mehrdad},
  journal={arXiv preprint arXiv:2507.11851},
  year={2025}
}

@article{hu2025fast,
  title={Fast and accurate causal parallel decoding using jacobi forcing},
  author={Hu, Lanxiang and Kou, Siqi and Fu, Yichao and Rajbhandari, Samyam and Rosing, Tajana and He, Yuxiong and Deng, Zhijie and Zhang, Hao},
  journal={arXiv preprint arXiv:2512.14681},
  year={2025}
}

@article{wu2025free,
  title={Free Draft-and-Verification: Toward Lossless Parallel Decoding for Diffusion Large Language Models},
  author={Wu, Shutong and Zhang, Jiawei},
  journal={arXiv preprint arXiv:2510.00294},
  year={2025}
}

@article{labs2025mercury,
  title={Mercury: Ultra-Fast Language Models Based on Diffusion},
  author={Labs, Inception and Khanna, Samar and Kharbanda, Siddhant and Li, Shufan and Varma, Harshit and Wang, Eric and Birnbaum, Sawyer and Luo, Ziyang and Miraoui, Yanis and Palrecha, Akash and others},
  journal={arXiv preprint arXiv:2506.17298},
  year={2025}
}

@inproceedings{
li2022diffusionlm,
title={Diffusion-{LM} Improves Controllable Text Generation},
author={Xiang Lisa Li and John Thickstun and Ishaan Gulrajani and Percy Liang and Tatsunori Hashimoto},
booktitle={Advances in Neural Information Processing Systems},
editor={Alice H. Oh and Alekh Agarwal and Danielle Belgrave and Kyunghyun Cho},
year={2022},
url={https://openreview.net/forum?id=3s9IrEsjLyk}
}

@misc{austin2023structureddenoisingdiffusionmodels,
      title={Structured Denoising Diffusion Models in Discrete State-Spaces}, 
      author={Jacob Austin and Daniel D. Johnson and Jonathan Ho and Daniel Tarlow and Rianne van den Berg},
      year={2023},
      eprint={2107.03006},
      archivePrefix={arXiv},
      primaryClass={cs.LG},
      url={https://arxiv.org/abs/2107.03006}, 
}

@misc{lou2024discretediffusionmodelingestimating,
      title={Discrete Diffusion Modeling by Estimating the Ratios of the Data Distribution}, 
      author={Aaron Lou and Chenlin Meng and Stefano Ermon},
      year={2024},
      eprint={2310.16834},
      archivePrefix={arXiv},
      primaryClass={stat.ML},
      url={https://arxiv.org/abs/2310.16834}, 
}

@misc{nie2025largelanguagediffusionmodels,
      title={Large Language Diffusion Models}, 
      author={Shen Nie and Fengqi Zhu and Zebin You and Xiaolu Zhang and Jingyang Ou and Jun Hu and Jun Zhou and Yankai Lin and Ji-Rong Wen and Chongxuan Li},
      year={2025},
      eprint={2502.09992},
      archivePrefix={arXiv},
      primaryClass={cs.CL},
      url={https://arxiv.org/abs/2502.09992}, 
}

@misc{gat2025setblockdecodinglanguage,
      title={Set Block Decoding is a Language Model Inference Accelerator}, 
      author={Itai Gat and Heli Ben-Hamu and Marton Havasi and Daniel Haziza and Jeremy Reizenstein and Gabriel Synnaeve and David Lopez-Paz and Brian Karrer and Yaron Lipman},
      year={2025},
      eprint={2509.04185},
      archivePrefix={arXiv},
      primaryClass={cs.LG},
      url={https://arxiv.org/abs/2509.04185}, 
}

@article{hu2026lightningrl,
  title={LightningRL: Breaking the Accuracy-Parallelism Trade-off of Block-wise dLLMs via Reinforcement Learning},
  author={Hu, Yanzhe and Jin, Yijie and Liu, Pengfei and Yu, Kai and Deng, Zhijie},
  journal={arXiv preprint arXiv:2603.13319},
  year={2026}
}

@article{cobbe2021training,
  title={Training verifiers to solve math word problems},
  author={Cobbe, Karl and Kosaraju, Vineet and Bavarian, Mohammad and Chen, Mark and Jun, Heewoo and Kaiser, Lukasz and Plappert, Matthias and Tworek, Jerry and Hilton, Jacob and Nakano, Reiichiro and others},
  journal={arXiv preprint arXiv:2110.14168},
  year={2021}
}

@article{clark2018think,
  title={Think you have solved question answering? try arc, the ai2 reasoning challenge},
  author={Clark, Peter and Cowhey, Isaac and Etzioni, Oren and Khot, Tushar and Sabharwal, Ashish and Schoenick, Carissa and Tafjord, Oyvind},
  journal={arXiv preprint arXiv:1803.05457},
  year={2018}
}

@article{hendrycks2020measuring,
  title={Measuring massive multitask language understanding},
  author={Hendrycks, Dan and Burns, Collin and Basart, Steven and Zou, Andy and Mazeika, Mantas and Song, Dawn and Steinhardt, Jacob},
  journal={arXiv preprint arXiv:2009.03300},
  year={2020}
}

@article{wang2024mmlu,
  title={Mmlu-pro: A more robust and challenging multi-task language understanding benchmark},
  author={Wang, Yubo and Ma, Xueguang and Zhang, Ge and Ni, Yuansheng and Chandra, Abhranil and Guo, Shiguang and Ren, Weiming and Arulraj, Aaran and He, Xuan and Jiang, Ziyan and others},
  journal={Advances in Neural Information Processing Systems},
  volume={37},
  pages={95266--95290},
  year={2024}
}

@inproceedings{rein2024gpqa,
  title={Gpqa: A graduate-level google-proof q\&a benchmark},
  author={Rein, David and Hou, Betty Li and Stickland, Asa Cooper and Petty, Jackson and Pang, Richard Yuanzhe and Dirani, Julien and Michael, Julian and Bowman, Samuel R},
  booktitle={First conference on language modeling},
  year={2024}
}

@article{hendrycks2021measuring,
  title={Measuring mathematical problem solving with the math dataset},
  author={Hendrycks, Dan and Burns, Collin and Kadavath, Saurav and Arora, Akul and Basart, Steven and Tang, Eric and Song, Dawn and Steinhardt, Jacob},
  journal={arXiv preprint arXiv:2103.03874},
  year={2021}
}

@inproceedings{liu2024mathbench,
  title={Mathbench: Evaluating the theory and application proficiency of llms with a hierarchical mathematics benchmark},
  author={Liu, Hongwei and Zheng, Zilong and Qiao, Yuxuan and Duan, Haodong and Fei, Zhiwei and Zhou, Fengzhe and Zhang, Wenwei and Zhang, Songyang and Lin, Dahua and Chen, Kai},
  booktitle={Findings of the Association for Computational Linguistics: ACL 2024},
  pages={6884--6915},
  year={2024}
}

@misc{aime,
  title        = {AIME Problems and Solutions},
  author       = {{AIME}},
  howpublished = {\url{https://artofproblemsolving.com/wiki/index.php/AIME_Problems_and_Solutions}},
}

@article{chen2021evaluating,
  title={Evaluating large language models trained on code},
  author={Chen, Mark and Tworek, Jerry and Jun, Heewoo and Yuan, Qiming and Pinto, Henrique Ponde De Oliveira and Kaplan, Jared and Edwards, Harri and Burda, Yuri and Joseph, Nicholas and Brockman, Greg and others},
  journal={arXiv preprint arXiv:2107.03374},
  year={2021}
}

@article{zhou2023instruction,
  title={Instruction-following evaluation for large language models},
  author={Zhou, Jeffrey and Lu, Tianjian and Mishra, Swaroop and Brahma, Siddhartha and Basu, Sujoy and Luan, Yi and Zhou, Denny and Hou, Le},
  journal={arXiv preprint arXiv:2311.07911},
  year={2023}
}

@article{jain2024livecodebench,
  title={Livecodebench: Holistic and contamination free evaluation of large language models for code},
  author={Jain, Naman and Han, King and Gu, Alex and Li, Wen-Ding and Yan, Fanjia and Zhang, Tianjun and Wang, Sida and Solar-Lezama, Armando and Sen, Koushik and Stoica, Ion},
  journal={arXiv preprint arXiv:2403.07974},
  year={2024}
}

@article{odena2021program,
  title={Program synthesis with large language models},
  author={Odena, Augustus and Sutton, Charles and Dohan, David Martin and Jiang, Ellen and Michalewski, Henryk and Austin, Jacob and Bosma, Maarten Paul and Nye, Maxwell and Terry, Michael and Le, Quoc V},
  journal={n/a, page n/a, n/a. N/a},
  year={2021}
}
